\documentclass{amsart}\usepackage[]{graphicx}\usepackage[]{color}
\makeatletter
\def\maxwidth{ %
  \ifdim\Gin@nat@width>\linewidth
    \linewidth
  \else
    \Gin@nat@width
  \fi
}
\makeatother

\definecolor{fgcolor}{rgb}{0.345, 0.345, 0.345}
\newcommand{\hlnum}[1]{\textcolor[rgb]{0.686,0.059,0.569}{#1}}%
\newcommand{\hlstr}[1]{\textcolor[rgb]{0.192,0.494,0.8}{#1}}%
\newcommand{\hlcom}[1]{\textcolor[rgb]{0.678,0.584,0.686}{\textit{#1}}}%
\newcommand{\hlopt}[1]{\textcolor[rgb]{0,0,0}{#1}}%
\newcommand{\hlstd}[1]{\textcolor[rgb]{0.345,0.345,0.345}{#1}}%
\newcommand{\hlkwb}[1]{\textcolor[rgb]{0.69,0.353,0.396}{#1}}%
\newcommand{\hlkwc}[1]{\textcolor[rgb]{0.333,0.667,0.333}{#1}}%
\newcommand{\hlkwd}[1]{\textcolor[rgb]{0.737,0.353,0.396}{\textbf{#1}}}%

\usepackage{framed}
\makeatletter
\newenvironment{kframe}{%
 \def\at@end@of@kframe{}%
 \ifinner\ifhmode%
  \def\at@end@of@kframe{\end{minipage}}%
  \begin{minipage}{\columnwidth}%
 \fi\fi%
 \def\FrameCommand##1{\hskip\@totalleftmargin \hskip-\fboxsep
 \colorbox{shadecolor}{##1}\hskip-\fboxsep
     \hskip-\linewidth \hskip-\@totalleftmargin \hskip\columnwidth}%
 \MakeFramed {\advance\hsize-\width
   \@totalleftmargin\z@ \linewidth\hsize
   \@setminipage}}%
 {\par\unskip\endMakeFramed%
 \at@end@of@kframe}
\makeatother

\definecolor{shadecolor}{rgb}{.97, .97, .97}
\definecolor{messagecolor}{rgb}{0, 0, 0}
\definecolor{warningcolor}{rgb}{1, 0, 1}
\definecolor{errorcolor}{rgb}{1, 0, 0}
\newenvironment{knitrout}{}{} 

\usepackage{alltt}
\usepackage{amsmath,amssymb,amsthm}
\usepackage[utf8]{inputenc}
\usepackage{soul}
\usepackage{natbib}
\usepackage{geometry}
\newcommand{\Rpackage}[1]{{\texttt{#1}}}

\newcommand{\R}{{\normalfont\textsf{R }}{}}

\newcommand{\code}[1]{\texttt{#1}}

\title{High-Dimensional Metrics in R}
\thanks{Version:  \today}
\author{Victor Chernozhukov, Christian Hansen, Martin Spindler}
\IfFileExists{upquote.sty}{\usepackage{upquote}}{}
\begin{document}
\begin{abstract}
The package High-dimensional Metrics (\Rpackage{hdm}) is an evolving collection of statistical methods for estimation and quantification of uncertainty in high-dimensional approximately sparse models. It focuses on providing confidence intervals and significance testing for (possibly many) low-dimensional subcomponents of the high-dimensional parameter vector. Efficient estimators and uniformly valid confidence intervals for regression coefficients on target variables (e.g., treatment or policy variable) in a high-dimensional approximately sparse regression model, for average treatment effect (ATE) and average treatment effect for the treated (ATET),  as well for extensions of these parameters to the endogenous setting are provided. 
Theory grounded, data-driven methods for selecting the penalization parameter in Lasso regressions under heteroscedastic and non-Gaussian errors are implemented. Moreover, joint/ simultaneous confidence intervals for regression coefficients of a high-dimensional sparse regression are implemented, including a joint significance test for Lasso regression. Data sets which have been used in the literature and might be useful for classroom demonstration and for testing new estimators are included.  \R and the package \Rpackage{hdm} are open-source software projects and can be freely downloaded from CRAN:
\texttt{http://cran.r-project.org}.
\end{abstract}

\maketitle

\pagestyle{myheadings}
\markboth{\sc High-Dimensional Metrics in \R}{\sc }

\tableofcontents
\section{Introduction}
Analysis of high-dimensional models, models in which the number of parameters to be estimated is large relative to the sample size, is becoming increasingly important. Such models arise naturally in modern data sets which have many measured characteristics available per individual observation as in, for example, population census data, scanner data, and text data.  Such models also arise naturally even in data with a small number of measured characteristics in situations where the exact functional form with which the observed variables enter the model is unknown and we create many technical variables, a dictionary, from the raw characteristics. Examples covered by this scenario include semiparametric models with nonparametric nuisance functions.  More generally, models with many parameters relative to the sample size often arise when attempting to model complex phenomena.

With increasing availability of such data sets in economics and other data science fields, new methods for analyzing those data have been developed. The \R package \Rpackage{hdm} contains implementations of recently developed methods for high-dimensional approximately sparse models, mainly relying on forms of lasso and post-lasso as well as related estimation and inference methods.  The methods are illustrated with econometric applications, but are also useful in other  disciplines such as medicine, biology, sociology or psychology. 

The methods which are implemented in this package are distinct from already available methods in other packages in the following four major ways: 
\begin{itemize}

\item[\textbf{1)}] First, we provide a version of Lasso regression that expressly handles and allows for non-Gaussian and heteroscedastic errors.

\item[\textbf{2)}] Second, we implement a theoretically grounded, data-driven choice of the penalty level $\lambda$ in the Lasso regressions. To underscore this choice,  we call the Lasso implementation in this package \textquotedblleft rigorous\textquotedblright Lasso (=\code{rlasso}). The prefix \textbf{r} in function names should underscore this. In high-dimensional settings cross-validation is very popular; but it lacks a theoretical justification for use in the present context and some theoretical proposals for the choice of $\lambda$ are often not feasible. Moreover, the theoretically grounded, data-driven choice redundantizes cross-validation which is time-consuming particularly in large data sets.  

\item[\textbf{3)}] Third, we provide efficient estimators and  uniformly valid confidence intervals for various low-dimensional causal/structural parameters  appearing in high-dimensional approximately sparse models. For example, we provide efficient estimators and uniformly valid confidence intervals for a regression coefficient on a target variable (e.g., a treatment or policy variable) in a high-dimensional sparse regression model. Target variable in this context means the object not interest, e.g. a prespecified regression coefficient. We also provide estimates and confidence intervals for average treatment effect (ATE) and average treatment effect for the treated (ATET),  as well  extensions of these parameters to the endogenous setting.

\item[\textbf{4)}] Fourth, joint/ simultaneous confidence intervals for estimated coefficients in a  high-dimensional approximately sparse models are provided, based on the methods and theory developed in \cite{BCK2014}. They proposed uniformly valid confidence regions for regressions coefficients in a high-dimensional sparse Z-estimation problems, which include median, mean, and many other regression problems as special cases. In this article we apply this method to the coefficients of a Lasso regression and highlight this method with an empirical example.

\end{itemize}

\section{How to get started}
\R is an open source software project and can be freely downloaded from the CRAN
website along with its associated documentation. The \R package \Rpackage{hdm} can be downloaded from \texttt{cran.r-project.org}. To install the hdm package from \R we simply type,

\begin{knitrout}
\definecolor{shadecolor}{rgb}{0.969, 0.969, 0.969}\color{fgcolor}\begin{kframe}
\begin{alltt}
\hlkwd{install.packages}\hlstd{(}\hlstr{"hdm"}\hlstd{)}
\end{alltt}
\end{kframe}
\end{knitrout}

\noindent
The most current version of the package (development version) is maintained at R-Forge and can installed by

\begin{knitrout}
\definecolor{shadecolor}{rgb}{0.969, 0.969, 0.969}\color{fgcolor}\begin{kframe}
\begin{alltt}
\hlkwd{install.packages}\hlstd{(}\hlstr{"hdm"}\hlstd{,} \hlkwc{repos} \hlstd{=} \hlstr{"http://R-Forge.R-project.org"}\hlstd{)}
\end{alltt}
\end{kframe}
\end{knitrout}

\noindent
Provided that your machine has a proper internet connection and you
have write permission in the appropriate system directories,
the installation of the package should proceed automatically.
Once the \texttt{hdm} package is installed, it can be loaded to the current \R session by the command,
\begin{knitrout}
\definecolor{shadecolor}{rgb}{0.969, 0.969, 0.969}\color{fgcolor}\begin{kframe}
\begin{alltt}
\hlkwd{library}\hlstd{(hdm)}
\end{alltt}
\end{kframe}
\end{knitrout}

Online help is available in two ways.  For example, 
you  can type:
\begin{knitrout}
\definecolor{shadecolor}{rgb}{0.969, 0.969, 0.969}\color{fgcolor}\begin{kframe}
\begin{alltt}
\hlkwd{help}\hlstd{(}\hlkwc{package} \hlstd{=} \hlstr{"hdm"}\hlstd{)}
\hlkwd{help}\hlstd{(rlasso)}
\end{alltt}
\end{kframe}
\end{knitrout}
The former command gives an overview over the available commands in the package, and
the latter gives detailed information about a specific command.

More generally one can initiate a web-browser help session with the command,
\begin{knitrout}
\definecolor{shadecolor}{rgb}{0.969, 0.969, 0.969}\color{fgcolor}\begin{kframe}
\begin{alltt}
\hlkwd{help.start}\hlstd{()}
\end{alltt}
\end{kframe}
\end{knitrout}
and navigate as desired.  The browser approach is better adapted to exploratory inquiries, while the command line approach is better suited to confirmatory ones.

A valuable feature of \R help files is that the examples used to illustrate commands are executable, so they can be pasted into an \R session or run as a group with
a command like,

\begin{knitrout}
\definecolor{shadecolor}{rgb}{0.969, 0.969, 0.969}\color{fgcolor}\begin{kframe}
\begin{alltt}
\hlkwd{example}\hlstd{(rlasso)}
\end{alltt}
\end{kframe}
\end{knitrout}


\section{Prediction using Approximate Sparsity}

\subsection{Prediction in Linear Models using Approximate Sparsity}

Consider high dimensional approximately sparse linear regression models. These models have a large number of regressors $p$, possibly much larger than the sample size $n$, but only a relatively small number $s =o(n)$ of these regressors are important for capturing accurately the main features of the regression function. The latter assumption makes it possible to estimate these models effectively by searching for approximately the right set of regressors.

The model reads \[ y_i = x_i' \beta_0  + \varepsilon_i, \quad \mathbb{E}[\varepsilon_i x_i]=0, \quad \beta_0 \in \mathbb{R}^p, 
\quad i=1,\ldots,n \]
where $y_i$ are observations of the response variable, $x_i=(x_{i,j}, \ldots, x_{i,p})$'s are observations of $p-$dimensional  regressors, and $\varepsilon_i$'s are centered disturbances, where possibly $p \gg n$.   Assume that the data sequence is
i.i.d. for the sake of exposition, although the framework covered is considerably more general. An important point is that the errors $\varepsilon_i$ may be non-Gaussian or heteroskedastic \citep{BCCH12}.

The model can be exactly sparse, namely
\[
\| \beta_0\|_0 \leq s = o(n),
\]
or approximately sparse, namely that the values of coefficients, sorted in decreasing
order, $(| \beta_0|_{(j)})_{j=1}^p$ obey,
\[
| \beta_0|_{(j)} \leq \mathsf{A} j^{-\mathsf{a}(\beta_0)},  \quad \mathsf{a}(\beta_0)>1/2, \quad j=1,...,p.
\]
An approximately sparse model can be well-approximated by an exactly sparse model
with sparsity index \[s \propto n^{1/(2 \mathsf{a}(\beta_0))}.\]

In order to get  theoretically justified performance guarantees,  we consider the Lasso 
estimator with data-driven penalty loadings: 
\[ \hat \beta = \arg \min_{\beta \in  \mathbb{R}^p} \mathbb{E}_n [(y_i - x_i' \beta)^2] + \frac{\lambda}{n} ||\hat{\Psi} \beta||_1 \]
where $||\beta||_1=\sum_{j=1}^p |\beta_j|$, $\hat{\Psi}=\mathrm{diag}(\hat{\psi}_1,\ldots,\hat{\psi}_p)$ is a diagonal matrix consisting of  penalty loadings, and $\mathbb{E}_n$ abbreviates the empirical average. The penalty loadings are chosen to insure basic equivariance of coefficient estimates to rescaling of $x_{i,j}$ and can also be chosen to address heteroskedasticity in model errors.   We discuss the choice of $\lambda$ and $\hat \Psi$ below.

Regularization by the $\ell_1$-norm naturally helps the Lasso estimator to avoid overfitting, but it also shrinks the fitted coefficients towards zero, causing a potentially significant bias. In order to remove some of this bias,  consider the Post-Lasso estimator that applies ordinary least squares to the model $\hat{T}$ selected by Lasso, formally, 
\[ \hat{T} = \text{support}(\hat{\beta}) = \{ j \in \{ 1, \ldots,p\}: \lvert \hat{\beta} \rvert >0 \}. \]
The Post-Lasso estimate is then defined as
\[ \tilde{\beta} \in \arg\min_{\beta \in \mathbb{R}^p}  \ \mathbb{E}_n \left( y_i - \sum_{j=1}^p x_{i,j} \beta_j \right) ^2: \beta_j=0 \quad \text{ if } \hat \beta_j = 0 , \quad \forall j. \]
In words, the estimator is ordinary least squares applied to the data after removing the regressors that were not selected by Lasso. The Post-Lasso estimator was introduced and analysed in \citet{BC-Postlasso}.

A crucial matter is the choice of the penalization parameter $\lambda$.   With the right choice
of the penalty level, Lasso and Post-Lasso estimators possess excellent performance guarantees: They both achieve the near-oracle rate for estimating the regression function, namely with probability $1- \gamma - o(1)$,
\[
\sqrt{\mathbb{E}_n [ (x_{i}'(\hat \beta - \beta_0))^2 ] } \lesssim \sqrt{(s/n) \log p}. 
\]

In high-dimensions setting, cross-validation is very popular in practice but lacks theoretical justification and so may not provide such a performance guarantee. In sharp contrast, the choice of the penalization parameter $\lambda$ in the Lasso and Post-Lasso methods in this package is theoretical grounded and feasible. Therefore we call the resulting method the \textquotedblleft rigorous\textquotedblright  Lasso method and hence add a prefix \textbf{r} to the function names.

\underline{In the case of homoscedasticity}, we set the penalty loadings $\hat{\psi}_j = \sqrt{\mathbb{E}_n x_{i,j}^2}$, which insures basic equivariance properties. There are two choices for penalty level $\lambda$:  the $X$-independent choice
and $X$-dependent choice.  In the $X$-independent choice we set the penalty level
to
\[ \lambda = 2c \sqrt{n} \hat{\sigma} \Phi^{-1}(1-\gamma/(2p)), \]
where $\Phi$ denotes the cumulative standard normal distribution, 
 $\hat \sigma$ is a preliminary estimate of $\sigma = \sqrt{\mathbb{E} \varepsilon^2}$,
and $c$ is a theoretical constant, which is set to $c=1.1$ by default for the Post-Lasso method and $c=.5$ for the Lasso method, and $\gamma$ is the probability level, which is set to $\gamma =.1$ by default.   The parameter $\gamma$ can be interpreted as the probability of mistakenly not removing $X$'s when all of them have zero coefficients.   In the X-dependent case the penalty level is calculated as
\[ \lambda = 2c \hat{\sigma} \Lambda(1-\gamma|X), \]
where
\[ \Lambda(1-\gamma|X)=(1-\gamma)-\text{quantile of}\quad n||\mathbb{E}_n[x_i e_i] ||_{\infty}|X,\]
where $X=[x_1, \ldots, x_n]'$ and $e_i$ are iid $N(0,1)$, generated independently from $X$; this quantity  is approximated by simulation. The $X$-independent penalty is more conservative than the $X$-dependent penalty. In particular the $X$-dependent penalty automatically adapts to highly correlated designs, using less aggressive penalization in this case \citet{BCH2011:InferenceGauss}.

\underline{In the case of heteroskedasticity}, the loadings are set to $\hat{\psi}_j=\sqrt{\mathbb{E}_n[x_{ij}^2 \hat \varepsilon_i^2]}$, where $\hat \varepsilon_i$ are preliminary estimates of the errors. The penalty level
can be $X$-independent \citep{BCCH12}:
\[ \lambda = 2c \sqrt{n} \Phi^{-1} (1-\gamma/(2p)), \]
or it can be X-dependent and estimated by a multiplier bootstrap procedure \citep{CCK:AOS13}:
\[ \lambda = c \times c_W(1-\gamma), \]
where $c_W(1-\gamma)$ is the $1-\gamma$-quantile of the random variable $W$, conditional on the data, where
\[ W:= n \max_{1 \leq j \leq p} |2\mathbb{E}_n [x_{ij} \hat{\varepsilon}_i e_i]|,\]
where $e_i$ are iid standard normal variables distributed independently from the data, and $ \hat{\varepsilon}_i$ denotes an estimate of the residuals.

Estimation proceeds by iteration.  The estimates of residuals $\hat \varepsilon_i$ are initialized by running least squares of $y_i$ on five regressors that are most correlated to $y_i$. This implies conservative starting values for $\lambda$ and the penalty loadings, and leads to the initial Lasso and Post-Lasso estimates, which are then further updated by iteration.  The resulting iterative procedure is fully justified in the theoretical literature.

\subsection{A Joint Significance Test for Lasso Regression}

A basic question frequently arising in empirical work is whether the Lasso regression has explanatory power, comparable to a F-test for the classical linear regression model. The construction of a joint significance test follows \citep{CCK:AOS13} (Appendix M), and can be described as: 

Based on the model $ y_i =a_0 + x_i' b_0 + \varepsilon_i$, the null hypothesis of joint statistical in-significance is  $b_0 = 0$. The alternative is that of the joint statistical significance: $b_0 \neq 0$. The null hypothesis implies that
 
$$ \mathbb{E} \left[ (y_i - a_0) x_i \right] = 0,$$
 
and restriction can be tested using the sup-score statistic:
 
$$S = \| \sqrt{n} \mathbb{E}_n \left[ (y_i - \hat a_0) x_i \right] \|_\infty,$$

where $\hat a_i =  \mathbb{E}_n [y_i]$.  The critical value for this statistic can be approximated  by the multiplier bootstrap procedure, which simulates the statistic:
 
$$ S^* = \| \sqrt{n} \mathbb{E}_n \left[ (y_i - \hat a_0) x_i g_i \right] \|_\infty,$$
 
where $g_i$'s are iid $N(0,1)$, conditional on the data. The $(1-\alpha)$-quantile of $S^*$ serves as the critical value, $c(1-\alpha)$. We reject the null if $S > c(1-\alpha)$ in favor of statistical significant, and we keep the null of non-significance otherwise. This test procedure is implemented in the package when calling the \texttt{summary}-method of \texttt{rlasso}-objects.

\subsection*{R implementation}  The function \texttt{rlasso} implements Lasso and post-Lasso, where the prefix ``r" signifies that these are theoretically rigorous versions of Lasso and post-Lasso. The default option is post-Lasso, \code{post=TRUE}.  This function returns an object of S3 class \code{rlasso} for which methods like \code{predict}, \code{print}, \code{summary} are provided.

 \code{lassoShooting.fit} is the computational algorithm that underlies the estimation procedure, which implements a version of the Shooting Lasso Algorithm \citep{Fu1998}. The user has several options for choosing the non-default options. Specifically, the user can decide if an unpenalized \code{intercept} should be included (\code{TRUE} by default). The option \code{penalty} of the function \code{rlasso} allows different choices for the penalization parameter and loadings. It allows for homoskedastic or heteroskedastic errors with default \code{homoscedastic = FALSE}. Moreover, the dependence structure of the design matrix might be taken into consideration for calculation of the penalization parameter with \code{X.dependent.lambda = TRUE}. In combination with these options, the option \code{lambda.start} allows the user to set a starting value for $\lambda$ for the different algorithms. Moreover, the user can provide her own fixed value for the penalty level -- instead of the data-driven methods discussed above --  by setting \code{homoscedastic = "none"} and  supplying the value via \code{lambda.start}.

The constants $c$ and $\gamma$ from above can be set in the option \code{penalty}.
The quantities $\hat{\varepsilon}$, $\hat{\Psi}$, $\hat{\sigma}$ are calculated in a iterative manner. The maximum number of iterations and the tolerance when the algorithms should  stop can be set with \code{control}.

The method \texttt{summary} of \texttt{rlasso}-objects displays additionally for model diagnosis the $R^2$ value, the adjusted $R^2$ with degrees of freedom equal to the number of selected parameters, and the sup-score statistic for joint significance -- described above -- with corresponding p-value.

\subsection*{Example}(Prediction Using Lasso and Post-Lasso) Consider generated data from a sparse linear model:
\begin{knitrout}
\definecolor{shadecolor}{rgb}{0.969, 0.969, 0.969}\color{fgcolor}\begin{kframe}
\begin{alltt}
\hlkwd{set.seed}\hlstd{(}\hlnum{12345}\hlstd{)}
\hlstd{n} \hlkwb{=} \hlnum{100}  \hlcom{#sample size}
\hlstd{p} \hlkwb{=} \hlnum{100}  \hlcom{# number of variables}
\hlstd{s} \hlkwb{=} \hlnum{3}  \hlcom{# nubmer of variables with non-zero coefficients}
\hlstd{X} \hlkwb{=} \hlkwd{matrix}\hlstd{(}\hlkwd{rnorm}\hlstd{(n} \hlopt{*} \hlstd{p),} \hlkwc{ncol} \hlstd{= p)}
\hlstd{beta} \hlkwb{=} \hlkwd{c}\hlstd{(}\hlkwd{rep}\hlstd{(}\hlnum{5}\hlstd{, s),} \hlkwd{rep}\hlstd{(}\hlnum{0}\hlstd{, p} \hlopt{-} \hlstd{s))}
\hlstd{Y} \hlkwb{=} \hlstd{X} \hlopt{%*%} \hlstd{beta} \hlopt{+} \hlkwd{rnorm}\hlstd{(n)}
\end{alltt}
\end{kframe}
\end{knitrout}
Next we estimate the model, print the results, and make in-sample and out-of sample predictions.
We can use  methods \code{print} and \code{summarize} to print the results, where
the option \code{all} can be set to \code{FALSE} to limit the print only to the non-zero coefficients.

\begin{knitrout}
\definecolor{shadecolor}{rgb}{0.969, 0.969, 0.969}\color{fgcolor}\begin{kframe}
\begin{alltt}
\hlstd{lasso.reg} \hlkwb{=} \hlkwd{rlasso}\hlstd{(Y} \hlopt{~} \hlstd{X,} \hlkwc{post} \hlstd{=} \hlnum{FALSE}\hlstd{)}  \hlcom{# use lasso, not-Post-lasso}
\hlcom{# lasso.reg = rlasso(X, Y, post=FALSE)}
\hlstd{sum.lasso} \hlkwb{<-} \hlkwd{summary}\hlstd{(lasso.reg,} \hlkwc{all} \hlstd{=} \hlnum{FALSE}\hlstd{)}  \hlcom{# can also do print(lasso.reg, all=FALSE)}
\end{alltt}
\begin{verbatim}
## 
## Call:
## rlasso.formula(formula = Y ~ X, post = FALSE)
## 
## Post-Lasso Estimation:  FALSE 
## 
## Total number of variables: 100
## Number of selected variables: 11 
## 
## Residuals: 
##      Min       1Q   Median       3Q      Max 
## -2.09008 -0.45801 -0.01237  0.50291  2.25098 
## 
##             Estimate
## (Intercept)    0.057
## 1              4.771
## 2              4.693
## 3              4.766
## 13            -0.045
## 15            -0.047
## 16            -0.005
## 19            -0.092
## 22            -0.027
## 40            -0.011
## 61             0.114
## 100           -0.025
## 
## Residual standard error: 0.8039
## Multiple R-squared:  0.9913
## Adjusted R-squared:  0.9902
## Joint significance test:
##  the sup score statistic for joint significance test is 64.02 with a p-value of     0
\end{verbatim}
\begin{alltt}
\hlstd{yhat.lasso} \hlkwb{=} \hlkwd{predict}\hlstd{(lasso.reg)}  \hlcom{#in-sample prediction}
\hlstd{Xnew} \hlkwb{=} \hlkwd{matrix}\hlstd{(}\hlkwd{rnorm}\hlstd{(n} \hlopt{*} \hlstd{p),} \hlkwc{ncol} \hlstd{= p)}  \hlcom{# new X}
\hlstd{Ynew} \hlkwb{=} \hlstd{Xnew} \hlopt{%*%} \hlstd{beta} \hlopt{+} \hlkwd{rnorm}\hlstd{(n)}  \hlcom{#new Y}
\hlstd{yhat.lasso.new} \hlkwb{=} \hlkwd{predict}\hlstd{(lasso.reg,} \hlkwc{newdata} \hlstd{= Xnew)}  \hlcom{#out-of-sample prediction}

\hlstd{post.lasso.reg} \hlkwb{=} \hlkwd{rlasso}\hlstd{(Y} \hlopt{~} \hlstd{X,} \hlkwc{post} \hlstd{=} \hlnum{TRUE}\hlstd{)}  \hlcom{#now use post-lasso}
\hlkwd{print}\hlstd{(post.lasso.reg,} \hlkwc{all} \hlstd{=} \hlnum{FALSE}\hlstd{)}  \hlcom{# or use  summary(post.lasso.reg, all=FALSE) }
\end{alltt}
\begin{verbatim}
## 
## Call:
## rlasso.formula(formula = Y ~ X, post = TRUE)
## 
## (Intercept)            1            2            3  
##      0.0341       4.9241       4.8579       4.9644
\end{verbatim}
\begin{alltt}
\hlstd{yhat.postlasso} \hlkwb{=} \hlkwd{predict}\hlstd{(post.lasso.reg)}  \hlcom{#in-sample prediction}
\hlstd{yhat.postlasso.new} \hlkwb{=} \hlkwd{predict}\hlstd{(post.lasso.reg,} \hlkwc{newdata} \hlstd{= Xnew)}  \hlcom{#out-of-sample prediction}
\hlstd{MAE} \hlkwb{<-} \hlkwd{apply}\hlstd{(}\hlkwd{cbind}\hlstd{(}\hlkwd{abs}\hlstd{(Ynew} \hlopt{-} \hlstd{yhat.lasso.new),} \hlkwd{abs}\hlstd{(Ynew} \hlopt{-} \hlstd{yhat.postlasso.new)),} \hlnum{2}\hlstd{,}
    \hlstd{mean)}
\hlkwd{names}\hlstd{(MAE)} \hlkwb{<-} \hlkwd{c}\hlstd{(}\hlstr{"lasso MAE"}\hlstd{,} \hlstr{"Post-lasso MAE"}\hlstd{)}
\hlkwd{print}\hlstd{(MAE,} \hlkwc{digits} \hlstd{=} \hlnum{2}\hlstd{)}  \hlcom{# MAE for Lasso and Post-Lasso}
\end{alltt}
\begin{verbatim}
##      lasso MAE Post-lasso MAE 
##           0.91           0.79
\end{verbatim}
\end{kframe}
\end{knitrout}

In the example above the sup-score statistic for overall significance is 64.02 with a pvalue of 0. This means that the null hypothesis is rejected on level $\alpha=0.05$ and the model seems to have explanatory power.

\section{Inference on Target Regression Coefficients}

Here we consider inference on the target coefficient $\alpha$ in the model:
$$
y_i = d_i \alpha_0 + x_i'\beta_0 + \epsilon_i,   \quad \mathbb{E} \epsilon_i (x_i', d_i')' =0.
$$
Here $d_i$ is a target regressor such as treatment, policy or other variable whose regression coefficient $\alpha_0$ we would like to learn \citep{BelloniChernozhukovHansen2011}.  If we are interested in coefficients of several or even many variables, we can simply  write the model in the above form treating each variable of interest as $d_i$ in turn and then applying the estimation and inference procedures described below.

We assume approximate sparsity for $x_i'\beta_0$ with sufficient speed of decay of the sorted components of $\beta_0$, namely $\mathsf{a}(\beta_0) >1$. This condition translates into having a sparsity index $s \ll \sqrt{n}$. In general $d_i$ is correlated to $x_i$, so $\alpha_0$ cannot be consistently estimated by the regression of $y_i$ on $d_i$. To keep track of the relationship of $d_i$ to $x_i$, write
$$
d_i = x_i'\pi^d_0 + \rho^d_i,  \quad \mathbb{E} \rho^d_i x_i = 0.
$$
To estimate $\alpha_0$, we also impose approximate sparsity on the regression function  $x_i'\pi^d_0$  with sufficient speed of decay of sorted components of $\pi^d_0$, namely $\mathsf{a}(\pi^d_0) > 1$.

\textbf{The Orthogonality Principle.} Note that we can not use naive estimates of $\alpha_0$ based simply on applying Lasso and Post-Lasso to the first equation. Such a strategy in general does not produce root-$n$ consistent and asymptotically normal estimators of $\alpha$, due to the possibility of large omitted variable bias resulting from estimating the nuisance function $x_i'\beta_0$ in high-dimensional setting. In order to overcome the omitted variable bias, we need to use orthogonalized estimating equations for $\alpha_0$. Specifically we seek to find a score $\psi(w_i, \alpha, \eta)$, where $w_i = (y_i,x_i')'$ and $\eta$ is the nuisance parameter, such that
$$
\mathbb{E} \psi(w_i, \alpha_0, \eta_0) = 0, \quad \frac{\partial}{\partial \eta} \mathbb{E} \psi(w_i, \alpha_0, \eta_0) = 0.
$$
The second equation is the orthogonality condition, which states that the equations are not sensitive to the first-order perturbations of the nuisance parameter $\eta$ near the true value.  The latter property allows estimation of these nuisance parameters $\eta_0$ by regularized estimators $\hat \eta$, where regularization is done via penalization or selection.  Without this property, regularization may have too much effect on the estimator of $\alpha_0$ for regular inference to proceed.  

The estimators $\hat \alpha$ of $\alpha_0$ solve the empirical analog
of the equation above,
$$
\mathbb{E}_n \psi(w_i, \hat \alpha, \hat \eta) = 0,
$$
where we have plugged in the estimator $\hat \eta$ for the nuisance parameter.  Due to the orthogonality property the estimator
is first-order equivalent to the infeasible estimator $\tilde \alpha$ solving 
$$
\mathbb{E}_n \psi(w_i, \tilde \alpha, \eta_0) = 0,
$$
where we use the true value of the nuisance parameter.  The equivalence holds in a variety of models under plausible conditions. The systematic development of the orthogonality condition for inference on low-dimensional parameters in modern high-dimensional settings is given in \citet{CHS2015}.

In turns out that in the linear model the orthogonal equations are closely connected to the classical ideas of partialling out.

\subsection{Intuition for the  Orthogonality Principle in Linear Models via Partialling Out}  One way to think about estimation of $\alpha_0$ is to think of the regression model:
$$
\rho^y_i = \alpha_0 \rho^d_i + \epsilon_i,  
$$
where $\rho^y_i$ is the residual that is left after partialling out the linear effect of $x_i$ from $y_i$ and $\rho^d_i$ is the residual that is left after partialling out the linear effect of $x_i$ from $d_i$, both done in the population.  Note that we have
$\mathbb{E} \rho^y_i  x_i =0$, i.e. $\rho^y_i = y_i - x_i'\pi^y_0$ where $x_i'\pi^y_0$ is the linear projection of $y_i$ on $x_i$.  After partialling out,  $\alpha_0$ is the population regression coefficient in the univariate regression of $\rho^y_i$ on $\rho^d_i$.  This is the Frisch-Waugh-Lovell theorem.   Thus, $\alpha=\alpha_0$ solves the population equation:
$$
\mathbb{E} (\rho^y_i - \alpha \rho^d_i)\rho^d_i = 0.
$$
The score associated to this equation is:
$$
\psi(w_i, \alpha, \eta) = (y_i - x_i'\pi^y) - \alpha (d_i - x_i'\pi^d))(d_i - x_i'\pi^d),  \quad \eta = (\pi^{y'}, \pi^{d'})', 
$$$$
\psi(w_i, \alpha_0, \eta_0) = (\rho^y_i - \alpha \rho^d_i)\rho^d_i, \quad \eta_0 = (\pi^{y'}_0, \pi^{d'}_0).
$$
It is straightforward to check that this score obeys the orthogonality principle; moreover, this score is the semi-parametrically efficient score for estimating the regression coefficient $\alpha_0$.

\underline{In low-dimensional settings}, the empirical version of the partialling out approach is simply another way to do the least squares.  Let's verify this in an example.
First, we generate some data
\begin{knitrout}
\definecolor{shadecolor}{rgb}{0.969, 0.969, 0.969}\color{fgcolor}\begin{kframe}
\begin{alltt}
\hlkwd{set.seed}\hlstd{(}\hlnum{1}\hlstd{)}
\hlstd{n} \hlkwb{=} \hlnum{5000}
\hlstd{p} \hlkwb{=} \hlnum{20}
\hlstd{X} \hlkwb{=} \hlkwd{matrix}\hlstd{(}\hlkwd{rnorm}\hlstd{(n} \hlopt{*} \hlstd{p),} \hlkwc{ncol} \hlstd{= p)}
\hlkwd{colnames}\hlstd{(X)} \hlkwb{=} \hlkwd{c}\hlstd{(}\hlstr{"d"}\hlstd{,} \hlkwd{paste}\hlstd{(}\hlstr{"x"}\hlstd{,} \hlnum{1}\hlopt{:}\hlnum{19}\hlstd{,} \hlkwc{sep} \hlstd{=} \hlstr{""}\hlstd{))}
\hlstd{xnames} \hlkwb{=} \hlkwd{colnames}\hlstd{(X)[}\hlopt{-}\hlnum{1}\hlstd{]}
\hlstd{beta} \hlkwb{=} \hlkwd{rep}\hlstd{(}\hlnum{1}\hlstd{,} \hlnum{20}\hlstd{)}
\hlstd{y} \hlkwb{=} \hlstd{X} \hlopt{%*%} \hlstd{beta} \hlopt{+} \hlkwd{rnorm}\hlstd{(n)}
\hlstd{dat} \hlkwb{=} \hlkwd{data.frame}\hlstd{(}\hlkwc{y} \hlstd{= y, X)}
\end{alltt}
\end{kframe}
\end{knitrout}
We can estimate $\alpha_0$ by running full least squares:
\begin{knitrout}
\definecolor{shadecolor}{rgb}{0.969, 0.969, 0.969}\color{fgcolor}\begin{kframe}
\begin{alltt}
\hlcom{# full fit}
\hlstd{fmla} \hlkwb{=} \hlkwd{as.formula}\hlstd{(}\hlkwd{paste}\hlstd{(}\hlstr{"y ~ "}\hlstd{,} \hlkwd{paste}\hlstd{(}\hlkwd{colnames}\hlstd{(X),} \hlkwc{collapse} \hlstd{=} \hlstr{"+"}\hlstd{)))}
\hlstd{full.fit} \hlkwb{=} \hlkwd{lm}\hlstd{(fmla,} \hlkwc{data} \hlstd{= dat)}
\hlkwd{summary}\hlstd{(full.fit)}\hlopt{$}\hlstd{coef[}\hlstr{"d"}\hlstd{,} \hlnum{1}\hlopt{:}\hlnum{2}\hlstd{]}
\end{alltt}
\begin{verbatim}
##   Estimate Std. Error 
## 0.97807455 0.01371225
\end{verbatim}
\end{kframe}
\end{knitrout}

Another way to estimate $\alpha_0$ is to first partial out 
the $x$-variables from $y_i$ and $d_i$, and run least squares
on the residuals: 
\begin{knitrout}
\definecolor{shadecolor}{rgb}{0.969, 0.969, 0.969}\color{fgcolor}\begin{kframe}
\begin{alltt}
\hlstd{fmla.y} \hlkwb{=} \hlkwd{as.formula}\hlstd{(}\hlkwd{paste}\hlstd{(}\hlstr{"y ~ "}\hlstd{,} \hlkwd{paste}\hlstd{(xnames,} \hlkwc{collapse} \hlstd{=} \hlstr{"+"}\hlstd{)))}
\hlstd{fmla.d} \hlkwb{=} \hlkwd{as.formula}\hlstd{(}\hlkwd{paste}\hlstd{(}\hlstr{"d ~ "}\hlstd{,} \hlkwd{paste}\hlstd{(xnames,} \hlkwc{collapse} \hlstd{=} \hlstr{"+"}\hlstd{)))}
\hlcom{# partial fit via ols}
\hlstd{rY} \hlkwb{=} \hlkwd{lm}\hlstd{(fmla.y,} \hlkwc{data} \hlstd{= dat)}\hlopt{$}\hlstd{res}
\hlstd{rD} \hlkwb{=} \hlkwd{lm}\hlstd{(fmla.d,} \hlkwc{data} \hlstd{= dat)}\hlopt{$}\hlstd{res}
\hlstd{partial.fit.ls} \hlkwb{=} \hlkwd{lm}\hlstd{(rY} \hlopt{~} \hlstd{rD)}
\hlkwd{summary}\hlstd{(partial.fit.ls)}\hlopt{$}\hlstd{coef[}\hlstr{"rD"}\hlstd{,} \hlnum{1}\hlopt{:}\hlnum{2}\hlstd{]}
\end{alltt}
\begin{verbatim}
##   Estimate Std. Error 
## 0.97807455 0.01368616
\end{verbatim}
\end{kframe}
\end{knitrout}
One can see that the estimates are identical, while standard errors are nearly identical. In fact, the standard errors are asymptotically equivalent due to the orthogonality property of the estimating equations associated with the partialling out approach.

\underline{In high-dimensional settings}, we can no longer rely on the full least-squares and instead may rely on Lasso or  Post-Lasso for partialling out. This amounts to using orthogonal estimating equations, where we estimate the nuisance parameters by Lasso or Post-Lasso.
Let's try this in the above example, using Post-Lasso for partialling out:
\begin{knitrout}
\definecolor{shadecolor}{rgb}{0.969, 0.969, 0.969}\color{fgcolor}\begin{kframe}
\begin{alltt}
\hlcom{# partial fit via post-lasso}
\hlstd{rY} \hlkwb{=} \hlkwd{rlasso}\hlstd{(fmla.y,} \hlkwc{data} \hlstd{= dat)}\hlopt{$}\hlstd{res}
\hlstd{rD} \hlkwb{=} \hlkwd{rlasso}\hlstd{(fmla.d,} \hlkwc{data} \hlstd{= dat)}\hlopt{$}\hlstd{res}
\hlstd{partial.fit.postlasso} \hlkwb{=} \hlkwd{lm}\hlstd{(rY} \hlopt{~} \hlstd{rD)}
\hlkwd{summary}\hlstd{(partial.fit.postlasso)}\hlopt{$}\hlstd{coef[}\hlstr{"rD"}\hlstd{,} \hlnum{1}\hlopt{:}\hlnum{2}\hlstd{]}
\end{alltt}
\begin{verbatim}
##   Estimate Std. Error 
## 0.97273870 0.01368677
\end{verbatim}
\end{kframe}
\end{knitrout}
We see that this estimate and standard errors are nearly identical to the previous estimates given above. In fact they are asymptotically equivalent to the previous estimates in the low-dimensional settings, with the advantage that, unlike the previous estimates, they will continue to be valid in the high-dimensional settings.

The orthogonal estimating equations method -- based on partialling out via Lasso or post-Lasso  -- is implemented by the function \texttt{rlassoEffect}, using \code{method= "partialling out"}:
\begin{knitrout}
\definecolor{shadecolor}{rgb}{0.969, 0.969, 0.969}\color{fgcolor}\begin{kframe}
\begin{alltt}
\hlstd{Eff} \hlkwb{=} \hlkwd{rlassoEffect}\hlstd{(X[,} \hlopt{-}\hlnum{1}\hlstd{], y, X[,} \hlnum{1}\hlstd{],} \hlkwc{method} \hlstd{=} \hlstr{"partialling out"}\hlstd{)}
\hlkwd{summary}\hlstd{(Eff)}\hlopt{$}\hlstd{coef[,} \hlnum{1}\hlopt{:}\hlnum{2}\hlstd{]}
\end{alltt}
\begin{verbatim}
##  Estimate. Std. Error 
## 0.97273870 0.01368677
\end{verbatim}
\end{kframe}
\end{knitrout}

Another orthogonal estimating equations method -- based on the double selection of covariates -- is implemented by the  the function \texttt{rlassoEffect}, using \code{method= "double selection"}.
Algorithmically the method is as follows:
\begin{enumerate}
  \item Select controls $x_{ij}$'s that predict $y_i$ by Lasso.
  \item Select controls $x_{ij}$'s that predict $d_i$ by Lasso.
  \item Run OLS of $y_i$ on $d_i$ and the union of controls selected in steps 1
  and 2.
\end{enumerate}
\begin{knitrout}
\definecolor{shadecolor}{rgb}{0.969, 0.969, 0.969}\color{fgcolor}\begin{kframe}
\begin{alltt}
\hlstd{Eff} \hlkwb{=} \hlkwd{rlassoEffect}\hlstd{(X[,} \hlopt{-}\hlnum{1}\hlstd{], y, X[,} \hlnum{1}\hlstd{],} \hlkwc{method} \hlstd{=} \hlstr{"double selection"}\hlstd{)}
\hlkwd{summary}\hlstd{(Eff)}\hlopt{$}\hlstd{coef[,} \hlnum{1}\hlopt{:}\hlnum{2}\hlstd{]}
\end{alltt}
\begin{verbatim}
##  Estimate. Std. Error 
## 0.97807455 0.01415624
\end{verbatim}
\end{kframe}
\end{knitrout}

\subsection{Inference: Confidence Intervals and Significance Testing}
The function \code{rlassoEffects} does inference -- namely construction of confidence intervals and significance testing -- for target variables. Those can be specified either by the variable names, an integer valued vector giving their position in \code{x} or by a logical vector indicating the variables for which inference should be conducted. It returns an object of S3 class \code{rlassoEffects} for which the methods \code{summary}, \code{print}, \code{confint}, and \code{plot} are provided. \code{rlassoEffects} is a wrap function for the function \code{rlassoEffect} which does inference for a single target regressor. The theoretical underpinning is given in \citet{BelloniChernozhukovHansen2011} and for a more general class of models in \citet{BCK2014}.
The function \code{rlassoEffects} might either be used in the form \code{rlassoEffects(x, y, index)} where \code{x} is a matrix, \code{y} denotes the outcome variable and \code{index} specifies the variables of \code{x} for which inference is conducted. This can done by an integer vector (postion of the variables), a logical vector or the name of the variables. An alternative usage is as \code{rlassoEffects(formula, data, I)} where \code{I} is a one-sided formula which specifies the variables for which is inference is conducted. For further details we refer to the help page of the function and the following examples where both methods for usage are shown.

Here is an example of the use.
\begin{knitrout}
\definecolor{shadecolor}{rgb}{0.969, 0.969, 0.969}\color{fgcolor}\begin{kframe}
\begin{alltt}
\hlkwd{set.seed}\hlstd{(}\hlnum{1}\hlstd{)}
\hlstd{n} \hlkwb{=} \hlnum{100}  \hlcom{#sample size}
\hlstd{p} \hlkwb{=} \hlnum{100}  \hlcom{# number of variables}
\hlstd{s} \hlkwb{=} \hlnum{3}  \hlcom{# nubmer of non-zero variables}
\hlstd{X} \hlkwb{=} \hlkwd{matrix}\hlstd{(}\hlkwd{rnorm}\hlstd{(n} \hlopt{*} \hlstd{p),} \hlkwc{ncol} \hlstd{= p)}
\hlkwd{colnames}\hlstd{(X)} \hlkwb{<-} \hlkwd{paste}\hlstd{(}\hlstr{"X"}\hlstd{,} \hlnum{1}\hlopt{:}\hlstd{p,} \hlkwc{sep} \hlstd{=} \hlstr{""}\hlstd{)}
\hlstd{beta} \hlkwb{=} \hlkwd{c}\hlstd{(}\hlkwd{rep}\hlstd{(}\hlnum{3}\hlstd{, s),} \hlkwd{rep}\hlstd{(}\hlnum{0}\hlstd{, p} \hlopt{-} \hlstd{s))}
\hlstd{y} \hlkwb{=} \hlnum{1} \hlopt{+} \hlstd{X} \hlopt{%*%} \hlstd{beta} \hlopt{+} \hlkwd{rnorm}\hlstd{(n)}
\hlstd{data} \hlkwb{=} \hlkwd{data.frame}\hlstd{(}\hlkwd{cbind}\hlstd{(y, X))}
\hlkwd{colnames}\hlstd{(data)[}\hlnum{1}\hlstd{]} \hlkwb{<-} \hlstr{"y"}
\hlstd{fm} \hlkwb{=} \hlkwd{paste}\hlstd{(}\hlstr{"y ~"}\hlstd{,} \hlkwd{paste}\hlstd{(}\hlkwd{colnames}\hlstd{(X),} \hlkwc{collapse} \hlstd{=} \hlstr{"+"}\hlstd{))}
\hlstd{fm} \hlkwb{=} \hlkwd{as.formula}\hlstd{(fm)}
\end{alltt}
\end{kframe}
\end{knitrout}
We can do inference on a set of variables of interest, e.g. the first,
second, third, and the fiftieth:
\begin{knitrout}
\definecolor{shadecolor}{rgb}{0.969, 0.969, 0.969}\color{fgcolor}\begin{kframe}
\begin{alltt}
\hlcom{# lasso.effect = rlassoEffects(X, y, index=c(1,2,3,50))}
\hlstd{lasso.effect} \hlkwb{=} \hlkwd{rlassoEffects}\hlstd{(fm,} \hlkwc{I} \hlstd{=} \hlopt{~}\hlstd{X1} \hlopt{+} \hlstd{X2} \hlopt{+} \hlstd{X3} \hlopt{+} \hlstd{X50,} \hlkwc{data} \hlstd{= data)}
\hlkwd{print}\hlstd{(lasso.effect)}
\end{alltt}
\begin{verbatim}
## 
## Call:
## rlassoEffects.formula(formula = fm, data = data, I = ~X1 + X2 + 
##     X3 + X50)
## 
## Coefficients:
##      X1       X2       X3      X50  
## 2.94448  3.04127  2.97540  0.07196
\end{verbatim}
\begin{alltt}
\hlkwd{summary}\hlstd{(lasso.effect)}
\end{alltt}
\begin{verbatim}
## [1] "Estimates and significance testing of the effect of target variables"
##     Estimate. Std. Error t value Pr(>|t|)    
## X1    2.94448    0.08815  33.404   <2e-16 ***
## X2    3.04127    0.08389  36.253   <2e-16 ***
## X3    2.97540    0.07804  38.127   <2e-16 ***
## X50   0.07196    0.07765   0.927    0.354    
## ---
## Signif. codes:  
## 0 '***' 0.001 '**' 0.01 '*' 0.05 '.' 0.1 ' ' 1
\end{verbatim}
\begin{alltt}
\hlkwd{confint}\hlstd{(lasso.effect)}
\end{alltt}
\begin{verbatim}
##           2.5 %    97.5 %
## X1   2.77171308 3.1172421
## X2   2.87685121 3.2056979
## X3   2.82244962 3.1283583
## X50 -0.08022708 0.2241377
\end{verbatim}
\end{kframe}
\end{knitrout}

The two methods are first-order equivalent in both low-dimensional and high-dimensional settings under regularity conditions.  Not surprisingly we see that in the numerical example of this section, the estimates and standard errors produced by the two methods are very close to each other.  

It is also possible to estimate joint confidence intervals. The method relies on a multiplier bootstrap method as described in \cite{BCK2014}. Joint confidence intervals can be invoked by setting the option \code{joint} to \code{TRUE} in the method \code{confint} for objects of class \code{rlassoEffects}. We will also demonstrate the application of joint confidence intervals in an empirical application in the next section.

\begin{knitrout}
\definecolor{shadecolor}{rgb}{0.969, 0.969, 0.969}\color{fgcolor}\begin{kframe}
\begin{alltt}
\hlkwd{confint}\hlstd{(lasso.effect,} \hlkwc{level} \hlstd{=} \hlnum{0.95}\hlstd{,} \hlkwc{joint} \hlstd{=} \hlnum{TRUE}\hlstd{)}
\end{alltt}
\begin{verbatim}
##          2.5 %    97.5 %
## X1   2.6470175 3.2419377
## X2   2.7608170 3.3217321
## X3   2.7115233 3.2392846
## X50 -0.1854959 0.3294065
\end{verbatim}
\end{kframe}
\end{knitrout}

Finally, we can also plot the estimated effects with their confidence intervals:
\begin{knitrout}
\definecolor{shadecolor}{rgb}{0.969, 0.969, 0.969}\color{fgcolor}\begin{kframe}
\begin{alltt}
\hlkwd{plot}\hlstd{(lasso.effect,} \hlkwc{main} \hlstd{=} \hlstr{"Confidence Intervals"}\hlstd{)}
\end{alltt}
\end{kframe}
\end{knitrout}

For logistic regression we provide the functions \code{rlassologit} and \code{rlassologitEffects}. Further information can be found in the help.

\subsection{Application: the effect of gender on wage}
In Labour Economics an important question is how the wage is related to the gender of the employed. We use US census data from the year 2012 to analyse the effect of gender and interaction effects of other variables with gender on wage jointly. The dependent variable is the logarithm of the wage, the target variable is \texttt{female} (in combination with other variables). All other variables denote some other socio-economic characteristics, e.g. marital status, education, and experience.  For a detailed description of the variables we refer to the help page.

First, we load and prepare the data.
\begin{knitrout}
\definecolor{shadecolor}{rgb}{0.969, 0.969, 0.969}\color{fgcolor}\begin{kframe}
\begin{alltt}
\hlkwd{library}\hlstd{(hdm)}
\hlkwd{data}\hlstd{(cps2012)}
\hlstd{X} \hlkwb{<-} \hlkwd{model.matrix}\hlstd{(}\hlopt{~-}\hlnum{1} \hlopt{+} \hlstd{female} \hlopt{+} \hlstd{female}\hlopt{:}\hlstd{(widowed} \hlopt{+} \hlstd{divorced} \hlopt{+} \hlstd{separated} \hlopt{+} \hlstd{nevermarried} \hlopt{+}
    \hlstd{hsd08} \hlopt{+} \hlstd{hsd911} \hlopt{+} \hlstd{hsg} \hlopt{+} \hlstd{cg} \hlopt{+} \hlstd{ad} \hlopt{+} \hlstd{mw} \hlopt{+} \hlstd{so} \hlopt{+} \hlstd{we} \hlopt{+} \hlstd{exp1} \hlopt{+} \hlstd{exp2} \hlopt{+} \hlstd{exp3)} \hlopt{+ +}\hlstd{(widowed} \hlopt{+}
    \hlstd{divorced} \hlopt{+} \hlstd{separated} \hlopt{+} \hlstd{nevermarried} \hlopt{+} \hlstd{hsd08} \hlopt{+} \hlstd{hsd911} \hlopt{+} \hlstd{hsg} \hlopt{+} \hlstd{cg} \hlopt{+} \hlstd{ad} \hlopt{+} \hlstd{mw} \hlopt{+} \hlstd{so} \hlopt{+}
    \hlstd{we} \hlopt{+} \hlstd{exp1} \hlopt{+} \hlstd{exp2} \hlopt{+} \hlstd{exp3)}\hlopt{^}\hlnum{2}\hlstd{,} \hlkwc{data} \hlstd{= cps2012)}
\hlkwd{dim}\hlstd{(X)}
\end{alltt}
\begin{verbatim}
## [1] 29217   136
\end{verbatim}
\begin{alltt}
\hlstd{X} \hlkwb{<-} \hlstd{X[,} \hlkwd{which}\hlstd{(}\hlkwd{apply}\hlstd{(X,} \hlnum{2}\hlstd{, var)} \hlopt{!=} \hlnum{0}\hlstd{)]}  \hlcom{# exclude all constant variables}
\hlkwd{dim}\hlstd{(X)}
\end{alltt}
\begin{verbatim}
## [1] 29217   116
\end{verbatim}
\begin{alltt}
\hlstd{index.gender} \hlkwb{<-} \hlkwd{grep}\hlstd{(}\hlstr{"female"}\hlstd{,} \hlkwd{colnames}\hlstd{(X))}
\hlstd{y} \hlkwb{<-} \hlstd{cps2012}\hlopt{$}\hlstd{lnw}
\end{alltt}
\end{kframe}
\end{knitrout}

The parameter estimates for the target parameters, i.e. all coefficients related to gender (i.e. by interaction with other variables) are calculated and summarized by the following commands

\begin{knitrout}
\definecolor{shadecolor}{rgb}{0.969, 0.969, 0.969}\color{fgcolor}\begin{kframe}
\begin{alltt}
\hlstd{effects.female} \hlkwb{<-} \hlkwd{rlassoEffects}\hlstd{(}\hlkwc{x} \hlstd{= X,} \hlkwc{y} \hlstd{= y,} \hlkwc{index} \hlstd{= index.gender)}
\hlkwd{summary}\hlstd{(effects.female)}
\end{alltt}
\begin{verbatim}
## [1] "Estimates and significance testing of the effect of target variables"
##                     Estimate. Std. Error t value Pr(>|t|)
## female              -0.154923   0.050162  -3.088 0.002012
## female:widowed       0.136095   0.090663   1.501 0.133325
## female:divorced      0.136939   0.022182   6.174 6.68e-10
## female:separated     0.023303   0.053212   0.438 0.661441
## female:nevermarried  0.186853   0.019942   9.370  < 2e-16
## female:hsd08         0.027810   0.120914   0.230 0.818092
## female:hsd911       -0.119335   0.051880  -2.300 0.021435
## female:hsg          -0.012890   0.019223  -0.671 0.502518
## female:cg            0.010139   0.018327   0.553 0.580114
## female:ad           -0.030464   0.021806  -1.397 0.162405
## female:mw           -0.001063   0.019192  -0.055 0.955811
## female:so           -0.008183   0.019357  -0.423 0.672468
## female:we           -0.004226   0.021168  -0.200 0.841760
## female:exp1          0.004935   0.007804   0.632 0.527139
## female:exp2         -0.159519   0.045300  -3.521 0.000429
## female:exp3          0.038451   0.007861   4.891 1.00e-06
##                        
## female              ** 
## female:widowed         
## female:divorced     ***
## female:separated       
## female:nevermarried ***
## female:hsd08           
## female:hsd911       *  
## female:hsg             
## female:cg              
## female:ad              
## female:mw              
## female:so              
## female:we              
## female:exp1            
## female:exp2         ***
## female:exp3         ***
## ---
## Signif. codes:  
## 0 '***' 0.001 '**' 0.01 '*' 0.05 '.' 0.1 ' ' 1
\end{verbatim}
\end{kframe}
\end{knitrout}

Finally, we estimate and plot confident intervals, first "pointwise" and then the joint confidence intervals.

\begin{knitrout}
\definecolor{shadecolor}{rgb}{0.969, 0.969, 0.969}\color{fgcolor}\begin{kframe}
\begin{alltt}
\hlstd{joint.CI} \hlkwb{<-} \hlkwd{confint}\hlstd{(effects.female,} \hlkwc{level} \hlstd{=} \hlnum{0.95}\hlstd{,} \hlkwc{joint} \hlstd{=} \hlnum{TRUE}\hlstd{)}
\hlstd{joint.CI}
\end{alltt}
\begin{verbatim}
##                            2.5 %      97.5 %
## female              -0.247543718 -0.06230284
## female:widowed      -0.043270502  0.31546147
## female:divorced      0.095621308  0.17825746
## female:separated    -0.069795172  0.11640070
## female:nevermarried  0.148558524  0.22514844
## female:hsd08        -0.239684558  0.29530518
## female:hsd911       -0.218862780 -0.01980730
## female:hsg          -0.047630251  0.02185069
## female:cg           -0.024316410  0.04459352
## female:ad           -0.073929927  0.01300244
## female:mw           -0.036731298  0.03460442
## female:so           -0.044444883  0.02807820
## female:we           -0.045052301  0.03660004
## female:exp1         -0.009313837  0.01918435
## female:exp2         -0.242381838 -0.07665682
## female:exp3          0.024087022  0.05281414
\end{verbatim}
\begin{alltt}
\hlcom{# plot(effects.female, joint=TRUE, level=0.95) # plot of the effects}
\end{alltt}
\end{kframe}
\end{knitrout}

This analysis allows a closer look how discrimination according to gender is related to other socio-economic variables.

As a side remark, the version 0.2 allows also now a formula interface for many functions including \texttt{rlassoEffects}. Hence, the analysis could also be done more compact as
\begin{knitrout}
\definecolor{shadecolor}{rgb}{0.969, 0.969, 0.969}\color{fgcolor}\begin{kframe}
\begin{alltt}
\hlstd{effects.female} \hlkwb{<-} \hlkwd{rlassoEffects}\hlstd{(lnw} \hlopt{~} \hlstd{female} \hlopt{+} \hlstd{female}\hlopt{:}\hlstd{(widowed} \hlopt{+} \hlstd{divorced} \hlopt{+} \hlstd{separated} \hlopt{+}
    \hlstd{nevermarried} \hlopt{+} \hlstd{hsd08} \hlopt{+} \hlstd{hsd911} \hlopt{+} \hlstd{hsg} \hlopt{+} \hlstd{cg} \hlopt{+} \hlstd{ad} \hlopt{+} \hlstd{mw} \hlopt{+} \hlstd{so} \hlopt{+} \hlstd{we} \hlopt{+} \hlstd{exp1} \hlopt{+} \hlstd{exp2} \hlopt{+}
    \hlstd{exp3)} \hlopt{+} \hlstd{(widowed} \hlopt{+} \hlstd{divorced} \hlopt{+} \hlstd{separated} \hlopt{+} \hlstd{nevermarried} \hlopt{+} \hlstd{hsd08} \hlopt{+} \hlstd{hsd911} \hlopt{+} \hlstd{hsg} \hlopt{+}
    \hlstd{cg} \hlopt{+} \hlstd{ad} \hlopt{+} \hlstd{mw} \hlopt{+} \hlstd{so} \hlopt{+} \hlstd{we} \hlopt{+} \hlstd{exp1} \hlopt{+} \hlstd{exp2} \hlopt{+} \hlstd{exp3)}\hlopt{^}\hlnum{2}\hlstd{,} \hlkwc{data} \hlstd{= cps2012,} \hlkwc{I} \hlstd{=} \hlopt{~}\hlstd{female} \hlopt{+}
    \hlstd{female}\hlopt{:}\hlstd{(widowed} \hlopt{+} \hlstd{divorced} \hlopt{+} \hlstd{separated} \hlopt{+} \hlstd{nevermarried} \hlopt{+} \hlstd{hsd08} \hlopt{+} \hlstd{hsd911} \hlopt{+} \hlstd{hsg} \hlopt{+}
        \hlstd{cg} \hlopt{+} \hlstd{ad} \hlopt{+} \hlstd{mw} \hlopt{+} \hlstd{so} \hlopt{+} \hlstd{we} \hlopt{+} \hlstd{exp1} \hlopt{+} \hlstd{exp2} \hlopt{+} \hlstd{exp3))}
\end{alltt}
\end{kframe}
\end{knitrout}
The one-sided option \texttt{I} gives the target variables for which inference is conducted.

\subsection{Application: Estimation of the treatment effect in a linear model with many confounding factors}

A part of empirical growth literature has focused on estimating the effect of an initial (lagged) level of GDP (Gross Domestic Product) per capita on the growth rates of GDP per capita. In particular, a key prediction from the classical Solow-Swan-Ramsey growth model is the hypothesis of convergence, which states that poorer countries should typically grow faster and therefore should tend to catch up with the richer countries, conditional on a set of institutional and societal characteristics. Covariates that describe such characteristics  include variables measuring education and science policies, strength of market institutions, trade openness, savings rates and others.

Thus, we are interested in a specification of the form:

\[
\label{GrowthEq}
y_i = \alpha_0 d_i+ \sum_{j=1}^p \beta_j x_{ij} + \varepsilon_i, \]
where $y_i$ is the growth rate of GDP over a specified decade in country $i$, $d_i$ is the log of the
initial level of GDP at the beginning of the specified period, and the $x_{ij}$'s form a
long list of country $i$'s characteristics at the beginning of the specified period. We
are interested in testing the hypothesis of convergence, namely that $\alpha_1 < 0$.
Given that in the \citet{BarroLee1994} data, the number
of covariates $p$ is large relative to the sample size $n$,
covariate selection becomes a crucial issue in this analysis.
We employ here the partialling out approach (as well as the double selection version) introduced in the previous section.

First, we load and prepare the data

\begin{knitrout}
\definecolor{shadecolor}{rgb}{0.969, 0.969, 0.969}\color{fgcolor}\begin{kframe}
\begin{alltt}
\hlkwd{data}\hlstd{(GrowthData)}
\hlkwd{dim}\hlstd{(GrowthData)}
\end{alltt}
\begin{verbatim}
## [1] 90 63
\end{verbatim}
\begin{alltt}
\hlstd{y} \hlkwb{=} \hlstd{GrowthData[,} \hlnum{1}\hlstd{,} \hlkwc{drop} \hlstd{= F]}
\hlstd{d} \hlkwb{=} \hlstd{GrowthData[,} \hlnum{3}\hlstd{,} \hlkwc{drop} \hlstd{= F]}
\hlstd{X} \hlkwb{=} \hlkwd{as.matrix}\hlstd{(GrowthData)[,} \hlopt{-}\hlkwd{c}\hlstd{(}\hlnum{1}\hlstd{,} \hlnum{2}\hlstd{,} \hlnum{3}\hlstd{)]}
\hlstd{varnames} \hlkwb{=} \hlkwd{colnames}\hlstd{(GrowthData)}
\end{alltt}
\end{kframe}
\end{knitrout}

Now we can estimate the effect of the initial GDP level. First, we estimate by OLS:
\begin{knitrout}
\definecolor{shadecolor}{rgb}{0.969, 0.969, 0.969}\color{fgcolor}\begin{kframe}
\begin{alltt}
\hlstd{xnames} \hlkwb{=} \hlstd{varnames[}\hlopt{-}\hlkwd{c}\hlstd{(}\hlnum{1}\hlstd{,} \hlnum{2}\hlstd{,} \hlnum{3}\hlstd{)]}  \hlcom{# names of X variables}
\hlstd{dandxnames} \hlkwb{=} \hlstd{varnames[}\hlopt{-}\hlkwd{c}\hlstd{(}\hlnum{1}\hlstd{,} \hlnum{2}\hlstd{)]}  \hlcom{# names of D and X variables}
\hlcom{# create formulas by pasting names (this saves typing times)}
\hlstd{fmla} \hlkwb{=} \hlkwd{as.formula}\hlstd{(}\hlkwd{paste}\hlstd{(}\hlstr{"Outcome ~ "}\hlstd{,} \hlkwd{paste}\hlstd{(dandxnames,} \hlkwc{collapse} \hlstd{=} \hlstr{"+"}\hlstd{)))}
\hlstd{ls.effect} \hlkwb{=} \hlkwd{lm}\hlstd{(fmla,} \hlkwc{data} \hlstd{= GrowthData)}
\end{alltt}
\end{kframe}
\end{knitrout}

Second, we estimate the effect by the partialling out by Post-Lasso: 
\begin{knitrout}
\definecolor{shadecolor}{rgb}{0.969, 0.969, 0.969}\color{fgcolor}\begin{kframe}
\begin{alltt}
\hlstd{dX} \hlkwb{=} \hlkwd{as.matrix}\hlstd{(}\hlkwd{cbind}\hlstd{(d, X))}
\hlstd{lasso.effect} \hlkwb{=} \hlkwd{rlassoEffect}\hlstd{(}\hlkwc{x} \hlstd{= X,} \hlkwc{y} \hlstd{= y,} \hlkwc{d} \hlstd{= d,} \hlkwc{method} \hlstd{=} \hlstr{"partialling out"}\hlstd{)}
\hlkwd{summary}\hlstd{(lasso.effect)}
\end{alltt}
\begin{verbatim}
## [1] "Estimates and significance testing of the effect of target variables"
##      Estimate. Std. Error t value Pr(>|t|)   
## [1,]  -0.04432    0.01532  -2.893  0.00381 **
## ---
## Signif. codes:  
## 0 '***' 0.001 '**' 0.01 '*' 0.05 '.' 0.1 ' ' 1
\end{verbatim}
\end{kframe}
\end{knitrout}

Third, we estimate the effect by the double selection method:
\begin{knitrout}
\definecolor{shadecolor}{rgb}{0.969, 0.969, 0.969}\color{fgcolor}\begin{kframe}
\begin{alltt}
\hlstd{dX} \hlkwb{=} \hlkwd{as.matrix}\hlstd{(}\hlkwd{cbind}\hlstd{(d, X))}
\hlstd{doublesel.effect} \hlkwb{=} \hlkwd{rlassoEffect}\hlstd{(}\hlkwc{x} \hlstd{= X,} \hlkwc{y} \hlstd{= y,} \hlkwc{d} \hlstd{= d,} \hlkwc{method} \hlstd{=} \hlstr{"double selection"}\hlstd{)}
\hlkwd{summary}\hlstd{(doublesel.effect)}
\end{alltt}
\begin{verbatim}
## [1] "Estimates and significance testing of the effect of target variables"
##          Estimate. Std. Error t value Pr(>|t|)   
## gdpsh465  -0.04432    0.01685  -2.631  0.00851 **
## ---
## Signif. codes:  
## 0 '***' 0.001 '**' 0.01 '*' 0.05 '.' 0.1 ' ' 1
\end{verbatim}
\end{kframe}
\end{knitrout}

We then collect results in a nice latex table:

\begin{knitrout}
\definecolor{shadecolor}{rgb}{0.969, 0.969, 0.969}\color{fgcolor}\begin{kframe}
\begin{alltt}
\hlkwd{library}\hlstd{(xtable)}
\hlstd{table} \hlkwb{=} \hlkwd{rbind}\hlstd{(}\hlkwd{summary}\hlstd{(ls.effect)}\hlopt{$}\hlstd{coef[}\hlstr{"gdpsh465"}\hlstd{,} \hlnum{1}\hlopt{:}\hlnum{2}\hlstd{],} \hlkwd{summary}\hlstd{(lasso.effect)}\hlopt{$}\hlstd{coef[,}
    \hlnum{1}\hlopt{:}\hlnum{2}\hlstd{],} \hlkwd{summary}\hlstd{(doublesel.effect)}\hlopt{$}\hlstd{coef[,} \hlnum{1}\hlopt{:}\hlnum{2}\hlstd{])}
\hlkwd{colnames}\hlstd{(table)} \hlkwb{=} \hlkwd{c}\hlstd{(}\hlstr{"Estimate"}\hlstd{,} \hlstr{"Std. Error"}\hlstd{)}  \hlcom{#names(summary(full.fit)$coef)[1:2]}
\hlkwd{rownames}\hlstd{(table)} \hlkwb{=} \hlkwd{c}\hlstd{(}\hlstr{"full reg via ols"}\hlstd{,} \hlstr{"partial reg
via post-lasso "}\hlstd{,} \hlstr{"partial reg via double selection"}\hlstd{)}
\hlstd{tab} \hlkwb{=} \hlkwd{xtable}\hlstd{(table,} \hlkwc{digits} \hlstd{=} \hlkwd{c}\hlstd{(}\hlnum{2}\hlstd{,} \hlnum{2}\hlstd{,} \hlnum{5}\hlstd{))}
\end{alltt}
\end{kframe}
\end{knitrout}

\begin{kframe}
\begin{alltt}
\hlstd{tab}
\end{alltt}
\end{kframe}
\begin{table}[ht]
\centering
\begin{tabular}{rrr}
  \hline
 & Estimate & Std. Error \\ 
  \hline
full reg via ols & -0.01 & 0.02989 \\ 
  partial reg
via post-lasso  & -0.04 & 0.01532 \\ 
  partial reg via double selection & -0.04 & 0.01685 \\ 
   \hline
\end{tabular}
\end{table}

We see that the OLS estimates are noisy, which is not surprising given that $p$ is comparable to $n$.  The partial regression approaches, based on Lasso selection of covariates in the two projection equations, in contrast yields much more precise estimates, which does support the hypothesis of conditional convergence.  We note that the partial regression approaches represent a special case of the orthogonal estimating equations approach.

\section{Instrumental Variable Estimation in a High-Dimensional Setting}

In many applied settings the researcher is interested in estimating the (structural) effect of a variable (treatment variable), but this variable is endogenous, i.e. correlated with the error term. In this case, instrumental variables (IV) methods are used for identification of the causal parameters.

We consider the linear instrumental variables model:
\begin{eqnarray}
y_i &=& \alpha_0 d_i + \gamma_0 x_i' + \varepsilon_i,\\
d_i &=& z_i' \Pi + \beta_0 x_i' + v_i,
\end{eqnarray}
where $\mathbb{E}[\varepsilon_i (x_i', z_i')]= 0$, $\mathbb{E}[v_i (x_i', z_i')]=0$, but $\mathbb{E}[\varepsilon_i v_i] \neq 0$ leading to endogeneity. In this setting $d_i$ is a scalar endogenous variable of interest, $z_i$ is a $p_z$-dimensional vector of instruments and $x_i$ is a $p_x$-dimensional vector of control variables.

In this section we present methods to estimate the effect $\alpha_0$ in a setting where either $x$ is high-dimensional or $z$ is high-dimensional. Instrumental variables estimation with very many instruments was analysed in \citet{BCCH12}, the extension with many instruments and many controls in \citet{CHS:ManyIVNote}.

\subsection{Estimation and Inference}

To get efficient estimators and uniformly valid confidence intervals for the structural parameters there are different strategies which are asymptotically equivalent where again orthogonalization (via partialling out) is a key concept.

In the case of the high-dimensional instrument $z_i$
and low-dimensional $x_i$. We predict the endogenous variable $d_i$ using (Post-)Lasso regression of $d_i$ on the instruments $z_i$ and $x_i$, forcing the inclusion of $x_i$. Then we compute the IV estimator (2SLS) $\hat \alpha$ of the parameter $\alpha_0$ using the predicted value $\hat d_i$ as instrument and using $x_i$'s as controls.   We then perform inference on $\alpha_0$ using $\hat \alpha$ and conventional heteroskedasticity robust standard errors.

In the case of the low-dimensional instrument $z_i$
and high-dimensional $x_i$, we first partial out the effect of $x_i$ from $d_i$, $y_i$, and $z_i$ by (Post-)Lasso. Second, we then use the residuals to compute the IV estimator (2SLS) $\hat \alpha$ of the parameter $\alpha_0$. We then perform inference on $\alpha_0$ using $\hat \alpha$ and conventional heteroskedasticity robust standard errors.

In the case of the high-dimensional instrument $z_i$
and high-dimensional $x_i$ the algorithm described in \citet{CHS:ManyIVNote} is adopted. 


\subsection*{R Implementation}
 The wrap function \code{rlassoIV} handles all of the previous cases. It has the options \code{select.X} and \code{select.Z} which implement selection of either covariates or instruments, both with default values set to \code{TRUE}. The class of the return object depends on the chosen options, but the methods \code{summary}, \code{print} and \code{confint} are available for all. The functions \code{rlassoSelectX} and \code{rlassoSelectZ} do selection on  $x$-variables only and  $z$-variables only. Selection on both is done in \code{rlassoIV}. All functions employ the option \code{post=TRUE} as default, which 
 uses post-Lasso for partialling out.  By setting \code{post=FALSE}
 we can employ Lasso instead of Post-Lasso. Finally, the package provides the standard function \code{tsls}, which implements the \textquotedblleft classical\textquotedblright\ two-stage least squares estimation.

\textbf{Fucntion usage} Both the family of \code{rlassoIV}-functions and the family of the functions for treatment effects , which are introduced in the next section, allow use with both formula-interface and by handing over the prepard model matrices. Hence the general pattern  for use with formula is \code{function(formula, data, ...)} where formula consists of two-parts and is a member of the class \code{Formula}. These formulas are of the pattern \code{y ~ d + x | x + z} where \code{y} is the outcome variable, \code{x} are exogenous variables, \code{d} endogenous varialbes (if several ones are allowed depends on the concrete function), and \code{z} denote the instrumental variables. A more primitive use of the functions is by simply hand over the required group of variables as matrices: \code{function(x=x, d=d, y=y, z=z)}. In some of the following examples both alternatives are displayed.   

\subsection{Application: Economic Development and Institutions}
Estimating the causal effect of institutions on output is complicated by the simultaneity between institutions and output: specifically, better institutions may
lead to higher incomes, but higher incomes may also lead to the development of
better institutions. To help overcome this simultaneity, \citet{acemoglu:colonial} use mortality rates for early European settlers as an instrument
for institution quality. The validity of this instrument hinges on the argument that
settlers set up better institutions in places where they are more likely to establish
long-term settlements, that where they are likely to settle for the long term is related
to settler mortality at the time of initial colonization, and that institutions are highly
persistent. The exclusion restriction for the instrumental variable is then motivated
by the argument that GDP, while persistent, is unlikely to be strongly influenced by
mortality in the previous century, or earlier, except through institutions.

In this application, we consider the problem of selecting
controls. The input raw controls are the Latitude and the continental dummies. The technical controls can include various polynomial transformations of the Latitude, possibly interacted with the continental dummies. Such flexible specifications of raw controls results in the high-dimensional $x$, with dimension comparable to the sample size.

First, we process the data

\begin{knitrout}
\definecolor{shadecolor}{rgb}{0.969, 0.969, 0.969}\color{fgcolor}\begin{kframe}
\begin{alltt}
\hlkwd{data}\hlstd{(AJR)}
\hlstd{y} \hlkwb{=} \hlstd{AJR}\hlopt{$}\hlstd{GDP}
\hlstd{d} \hlkwb{=} \hlstd{AJR}\hlopt{$}\hlstd{Exprop}
\hlstd{z} \hlkwb{=} \hlstd{AJR}\hlopt{$}\hlstd{logMort}
\hlstd{x} \hlkwb{=} \hlkwd{model.matrix}\hlstd{(}\hlopt{~-}\hlnum{1} \hlopt{+} \hlstd{(Latitude} \hlopt{+} \hlstd{Latitude2} \hlopt{+} \hlstd{Africa} \hlopt{+} \hlstd{Asia} \hlopt{+} \hlstd{Namer} \hlopt{+} \hlstd{Samer)}\hlopt{^}\hlnum{2}\hlstd{,}
    \hlkwc{data} \hlstd{= AJR)}
\hlkwd{dim}\hlstd{(x)}
\end{alltt}
\begin{verbatim}
## [1] 64 21
\end{verbatim}
\end{kframe}
\end{knitrout}

Then we estimate an IV model with selection on the $X$
\begin{knitrout}
\definecolor{shadecolor}{rgb}{0.969, 0.969, 0.969}\color{fgcolor}\begin{kframe}
\begin{alltt}
\hlcom{# AJR.Xselect = rlassoIV(x=x, d=d, y=y, z=z, select.X=TRUE, select.Z=FALSE)}
\hlstd{AJR.Xselect} \hlkwb{=} \hlkwd{rlassoIV}\hlstd{(GDP} \hlopt{~} \hlstd{Exprop} \hlopt{+} \hlstd{(Latitude} \hlopt{+} \hlstd{Latitude2} \hlopt{+} \hlstd{Africa} \hlopt{+} \hlstd{Asia} \hlopt{+} \hlstd{Namer} \hlopt{+}
    \hlstd{Samer)}\hlopt{^}\hlnum{2} \hlopt{|} \hlstd{logMort} \hlopt{+} \hlstd{(Latitude} \hlopt{+} \hlstd{Latitude2} \hlopt{+} \hlstd{Africa} \hlopt{+} \hlstd{Asia} \hlopt{+} \hlstd{Namer} \hlopt{+} \hlstd{Samer)}\hlopt{^}\hlnum{2}\hlstd{,}
    \hlkwc{data} \hlstd{= AJR,} \hlkwc{select.X} \hlstd{=} \hlnum{TRUE}\hlstd{,} \hlkwc{select.Z} \hlstd{=} \hlnum{FALSE}\hlstd{)}
\hlkwd{summary}\hlstd{(AJR.Xselect)}
\end{alltt}
\begin{verbatim}
## [1] "Estimation and significance testing of the effect of target variables in the IV regression model"
##        coeff.    se. t-value p-value   
## Exprop 0.8792 0.2975   2.956 0.00312 **
## ---
## Signif. codes:  
## 0 '***' 0.001 '**' 0.01 '*' 0.05 '.' 0.1 ' ' 1
\end{verbatim}
\begin{alltt}
\hlkwd{confint}\hlstd{(AJR.Xselect)}
\end{alltt}
\begin{verbatim}
##            2.5 %   97.5 %
## Exprop 0.2961694 1.462225
\end{verbatim}
\end{kframe}
\end{knitrout}

It is interesting to understand what the procedure above is doing.
 In essence, it partials out $x_i$ from $y_i$, $d_i$ and $z_i$ using Post-Lasso and applies the 2SLS to the residual quantities. 

Let us investigate partialling out in more detail in this example.  We can first try to use OLS for partialling out:
\begin{knitrout}
\definecolor{shadecolor}{rgb}{0.969, 0.969, 0.969}\color{fgcolor}\begin{kframe}
\begin{alltt}
\hlcom{# parialling out by linear model}
\hlstd{fmla.y} \hlkwb{=} \hlstd{GDP} \hlopt{~} \hlstd{(Latitude} \hlopt{+} \hlstd{Latitude2} \hlopt{+} \hlstd{Africa} \hlopt{+} \hlstd{Asia} \hlopt{+} \hlstd{Namer} \hlopt{+} \hlstd{Samer)}\hlopt{^}\hlnum{2}
\hlstd{fmla.d} \hlkwb{=} \hlstd{Exprop} \hlopt{~} \hlstd{(Latitude} \hlopt{+} \hlstd{Latitude2} \hlopt{+} \hlstd{Africa} \hlopt{+} \hlstd{Asia} \hlopt{+} \hlstd{Namer} \hlopt{+} \hlstd{Samer)}\hlopt{^}\hlnum{2}
\hlstd{fmla.z} \hlkwb{=} \hlstd{logMort} \hlopt{~} \hlstd{(Latitude} \hlopt{+} \hlstd{Latitude2} \hlopt{+} \hlstd{Africa} \hlopt{+} \hlstd{Asia} \hlopt{+} \hlstd{Namer} \hlopt{+} \hlstd{Samer)}\hlopt{^}\hlnum{2}
\hlstd{rY} \hlkwb{=} \hlkwd{lm}\hlstd{(fmla.y,} \hlkwc{data} \hlstd{= AJR)}\hlopt{$}\hlstd{res}
\hlstd{rD} \hlkwb{=} \hlkwd{lm}\hlstd{(fmla.d,} \hlkwc{data} \hlstd{= AJR)}\hlopt{$}\hlstd{res}
\hlstd{rZ} \hlkwb{=} \hlkwd{lm}\hlstd{(fmla.z,} \hlkwc{data} \hlstd{= AJR)}\hlopt{$}\hlstd{res}
\hlcom{# ivfit.lm = tsls(y=rY,d=rD, x=NULL, z=rZ, intercept=FALSE)}
\hlstd{ivfit.lm} \hlkwb{=} \hlkwd{tsls}\hlstd{(rY} \hlopt{~} \hlstd{rD} \hlopt{|} \hlstd{rZ,} \hlkwc{intercept} \hlstd{=} \hlnum{FALSE}\hlstd{)}
\hlkwd{print}\hlstd{(}\hlkwd{cbind}\hlstd{(ivfit.lm}\hlopt{$}\hlstd{coef, ivfit.lm}\hlopt{$}\hlstd{se),} \hlkwc{digits} \hlstd{=} \hlnum{3}\hlstd{)}
\end{alltt}
\begin{verbatim}
##    [,1] [,2]
## rD 1.27 1.73
\end{verbatim}
\end{kframe}
\end{knitrout}
We see that the estimates exhibit large standard errors.  The imprecision is expected because dimension of $x$ is quite large, comparable to the sample size.

Next, we replace the OLS operator by post-Lasso for partialling out
\begin{knitrout}
\definecolor{shadecolor}{rgb}{0.969, 0.969, 0.969}\color{fgcolor}\begin{kframe}
\begin{alltt}
\hlcom{# parialling out by lasso}
\hlstd{rY} \hlkwb{=} \hlkwd{rlasso}\hlstd{(fmla.y,} \hlkwc{data} \hlstd{= AJR)}\hlopt{$}\hlstd{res}
\hlstd{rD} \hlkwb{=} \hlkwd{rlasso}\hlstd{(fmla.d,} \hlkwc{data} \hlstd{= AJR)}\hlopt{$}\hlstd{res}
\hlstd{rZ} \hlkwb{=} \hlkwd{rlasso}\hlstd{(fmla.z,} \hlkwc{data} \hlstd{= AJR)}\hlopt{$}\hlstd{res}
\hlcom{# ivfit.lasso = tsls(y=rY,d=rD, x=NULL, z=rZ, intercept=FALSE)}
\hlstd{ivfit.lasso} \hlkwb{=} \hlkwd{tsls}\hlstd{(rY} \hlopt{~} \hlstd{rD} \hlopt{|} \hlstd{rZ,} \hlkwc{intercept} \hlstd{=} \hlnum{FALSE}\hlstd{)}
\hlkwd{print}\hlstd{(}\hlkwd{cbind}\hlstd{(ivfit.lasso}\hlopt{$}\hlstd{coef, ivfit.lasso}\hlopt{$}\hlstd{se),} \hlkwc{digits} \hlstd{=} \hlnum{3}\hlstd{)}
\end{alltt}
\begin{verbatim}
##     [,1]  [,2]
## rD 0.879 0.297
\end{verbatim}
\end{kframe}
\end{knitrout}
This is exactly what command \code{rlassoIV} is doing by calling the command \code{rlassoSelectX}, so the numbers we see are identical
to those reported first.  In comparison to OLS results,
we see precision is improved by doing selection on the exogenous variables.

\subsection{Application: Impact of Eminent Domain Decisions on Economic Outcomes}
Here we investigate the effect of pro-plaintiff decisions in cases of
eminent domain (government's takings of private property) on economic outcomes.  The analysis of the effects of such decisions is complicated by the possible endogeneity
between judicial decisions and potential economic outcomes. To address the potential endogeneity, we employ an instrumental
variables strategy based on the random assignment of judges to the federal appellate panels that make the decisions. Because judges are randomly assigned to three-judge panels, the exact identity of the judges and their demographics are randomly assigned conditional on the distribution of characteristics of federal circuit court judges in a given circuit-year. Under this random assignment, the
characteristics of judges serving on federal appellate panels can only be related to property prices through the judges' decisions; thus the judge's characteristics will plausibly satisfy the instrumental variable exclusion restriction. For further information on this application and the data set we refer to \citet{chen:yeh:takings} and \citet{BCCH12}.

First, we load the data an construct the matrices with the controls (x), instruments (z), outcome (y), and treatment variables (d). Here we consider regional GDP as the outcome variable. 
\begin{knitrout}
\definecolor{shadecolor}{rgb}{0.969, 0.969, 0.969}\color{fgcolor}\begin{kframe}
\begin{alltt}
\hlkwd{data}\hlstd{(EminentDomain)}
\hlstd{z} \hlkwb{<-} \hlkwd{as.matrix}\hlstd{(EminentDomain}\hlopt{$}\hlstd{logGDP}\hlopt{$}\hlstd{z)}
\hlstd{x} \hlkwb{<-} \hlkwd{as.matrix}\hlstd{(EminentDomain}\hlopt{$}\hlstd{logGDP}\hlopt{$}\hlstd{x)}
\hlstd{y} \hlkwb{<-} \hlstd{EminentDomain}\hlopt{$}\hlstd{logGDP}\hlopt{$}\hlstd{y}
\hlstd{d} \hlkwb{<-} \hlstd{EminentDomain}\hlopt{$}\hlstd{logGDP}\hlopt{$}\hlstd{d}
\hlstd{x} \hlkwb{<-} \hlstd{x[,} \hlkwd{apply}\hlstd{(x,} \hlnum{2}\hlstd{, mean,} \hlkwc{na.rm} \hlstd{=} \hlnum{TRUE}\hlstd{)} \hlopt{>} \hlnum{0.05}\hlstd{]}  \hlcom{#}
\hlstd{z} \hlkwb{<-} \hlstd{z[,} \hlkwd{apply}\hlstd{(z,} \hlnum{2}\hlstd{, mean,} \hlkwc{na.rm} \hlstd{=} \hlnum{TRUE}\hlstd{)} \hlopt{>} \hlnum{0.05}\hlstd{]}  \hlcom{# }
\end{alltt}
\end{kframe}
\end{knitrout}
As mentioned above, $y$ is the economic outcome, the logarithm of the GDP, $d$  the number of pro plaintiff appellate takings decisions in federal circuit court $c$ and year $t$, $x$ is a matrix with control variables, and $z$ is the matrix with instruments. Here we consider socio-economic and demographic characteristics of the judges as instruments.

First, we estimate the effect of the treatment variable by simple OLS and 2SLS using two instruments:
\begin{knitrout}
\definecolor{shadecolor}{rgb}{0.969, 0.969, 0.969}\color{fgcolor}\begin{kframe}
\begin{alltt}
\hlstd{ED.ols} \hlkwb{=} \hlkwd{lm}\hlstd{(y} \hlopt{~} \hlkwd{cbind}\hlstd{(d, x))}
\hlstd{ED.2sls} \hlkwb{=} \hlkwd{tsls}\hlstd{(}\hlkwc{y} \hlstd{= y,} \hlkwc{d} \hlstd{= d,} \hlkwc{x} \hlstd{= x,} \hlkwc{z} \hlstd{= z[,} \hlnum{1}\hlopt{:}\hlnum{2}\hlstd{],} \hlkwc{intercept} \hlstd{=} \hlnum{FALSE}\hlstd{)}
\end{alltt}
\end{kframe}
\end{knitrout}

Next, we estimate the model with selection on the instruments. 
\begin{knitrout}
\definecolor{shadecolor}{rgb}{0.969, 0.969, 0.969}\color{fgcolor}\begin{kframe}
\begin{alltt}
\hlstd{lasso.IV.Z} \hlkwb{=} \hlkwd{rlassoIV}\hlstd{(}\hlkwc{x} \hlstd{= x,} \hlkwc{d} \hlstd{= d,} \hlkwc{y} \hlstd{= y,} \hlkwc{z} \hlstd{= z,} \hlkwc{select.X} \hlstd{=} \hlnum{FALSE}\hlstd{,} \hlkwc{select.Z} \hlstd{=} \hlnum{TRUE}\hlstd{)}
\hlcom{# or lasso.IV.Z = rlassoIVselectZt(x=X, d=d, y=y, z=z)}
\hlkwd{summary}\hlstd{(lasso.IV.Z)}
\end{alltt}
\begin{verbatim}
## [1] "Estimates and significance testing of the effect of target variables in the IV regression model"
##    coeff.    se. t-value p-value
## d1 0.4146 0.2902   1.428   0.153
\end{verbatim}
\begin{alltt}
\hlkwd{confint}\hlstd{(lasso.IV.Z)}
\end{alltt}
\begin{verbatim}
##         2.5 %    97.5 %
## d1 -0.1542764 0.9834796
\end{verbatim}
\end{kframe}
\end{knitrout}

Finally, we do selection on both the $x$ and $z$ variables.
\begin{knitrout}
\definecolor{shadecolor}{rgb}{0.969, 0.969, 0.969}\color{fgcolor}\begin{kframe}
\begin{alltt}
\hlstd{lasso.IV.XZ} \hlkwb{=} \hlkwd{rlassoIV}\hlstd{(}\hlkwc{x} \hlstd{= x,} \hlkwc{d} \hlstd{= d,} \hlkwc{y} \hlstd{= y,} \hlkwc{z} \hlstd{= z,} \hlkwc{select.X} \hlstd{=} \hlnum{TRUE}\hlstd{,} \hlkwc{select.Z} \hlstd{=} \hlnum{TRUE}\hlstd{)}
\hlkwd{summary}\hlstd{(lasso.IV.XZ)}
\end{alltt}
\begin{verbatim}
## Estimates and Significance Testing of the effect of target variables in the IV regression model 
##      coeff.      se. t-value p-value
## d1 -0.03078  0.16077  -0.191   0.848
\end{verbatim}
\begin{alltt}
\hlkwd{confint}\hlstd{(lasso.IV.XZ)}
\end{alltt}
\begin{verbatim}
##         2.5 %    97.5 %
## d1 -0.3458816 0.2843283
\end{verbatim}
\end{kframe}
\end{knitrout}

Comparing the results we see, that the OLS estimates indicate that the influence  of pro plaintiff appellate takings decisions in federal circuit court is significantly positive, but the 2SLS estimates which account for the potential endogeneity render the results insignificant. Employing selection on the instruments yields similar results. Doing selection on both the $x$- and $z$-variables results in extremely imprecise estimates.

Finally, we compare all results

\begin{knitrout}
\definecolor{shadecolor}{rgb}{0.969, 0.969, 0.969}\color{fgcolor}\begin{kframe}
\begin{alltt}
\hlkwd{library}\hlstd{(xtable)}
\hlstd{table} \hlkwb{=} \hlkwd{matrix}\hlstd{(}\hlnum{0}\hlstd{,} \hlnum{4}\hlstd{,} \hlnum{2}\hlstd{)}
\hlstd{table[}\hlnum{1}\hlstd{, ]} \hlkwb{=} \hlkwd{summary}\hlstd{(ED.ols)}\hlopt{$}\hlstd{coef[}\hlnum{2}\hlstd{,} \hlnum{1}\hlopt{:}\hlnum{2}\hlstd{]}
\hlstd{table[}\hlnum{2}\hlstd{, ]} \hlkwb{=} \hlkwd{cbind}\hlstd{(ED.2sls}\hlopt{$}\hlstd{coef[}\hlnum{1}\hlstd{], ED.2sls}\hlopt{$}\hlstd{se[}\hlnum{1}\hlstd{])}
\hlstd{table[}\hlnum{3}\hlstd{, ]} \hlkwb{=} \hlkwd{summary}\hlstd{(lasso.IV.Z)[,} \hlnum{1}\hlopt{:}\hlnum{2}\hlstd{]}
\end{alltt}
\begin{verbatim}
## [1] "Estimates and significance testing of the effect of target variables in the IV regression model"
##    coeff.    se. t-value p-value
## d1 0.4146 0.2902   1.428   0.153
\end{verbatim}
\begin{alltt}
\hlstd{table[}\hlnum{4}\hlstd{, ]} \hlkwb{=} \hlkwd{summary}\hlstd{(lasso.IV.XZ)[,} \hlnum{1}\hlopt{:}\hlnum{2}\hlstd{]}
\end{alltt}
\begin{verbatim}
## Estimates and Significance Testing of the effect of target variables in the IV regression model 
##      coeff.      se. t-value p-value
## d1 -0.03078  0.16077  -0.191   0.848
\end{verbatim}
\begin{alltt}
\hlkwd{colnames}\hlstd{(table)} \hlkwb{=} \hlkwd{c}\hlstd{(}\hlstr{"Estimate"}\hlstd{,} \hlstr{"Std. Error"}\hlstd{)}
\hlkwd{rownames}\hlstd{(table)} \hlkwb{=} \hlkwd{c}\hlstd{(}\hlstr{"ols regression"}\hlstd{,} \hlstr{"IV estimation "}\hlstd{,} \hlstr{"selection on Z"}\hlstd{,} \hlstr{"selection on X and Z"}\hlstd{)}
\hlstd{tab} \hlkwb{=} \hlkwd{xtable}\hlstd{(table,} \hlkwc{digits} \hlstd{=} \hlkwd{c}\hlstd{(}\hlnum{2}\hlstd{,} \hlnum{2}\hlstd{,} \hlnum{7}\hlstd{))}
\end{alltt}
\end{kframe}
\end{knitrout}

\begin{kframe}
\begin{alltt}
\hlstd{tab}
\end{alltt}
\end{kframe}
\begin{table}[ht]
\centering
\begin{tabular}{rrr}
  \hline
 & Estimate & Std. Error \\ 
  \hline
ols regression & 0.01 & 0.0098659 \\ 
  IV estimation  & -0.01 & 0.0337664 \\ 
  selection on Z & 0.41 & 0.2902492 \\ 
  selection on X and Z & -0.03 & 0.1607708 \\ 
   \hline
\end{tabular}
\end{table}

\section{Inference on Treatment Effects in a High-Dimensional Setting}

In this section, we consider estimation and inference on treatment effects when the treatment variable $d$ enters non-separably in determination of the outcomes. This case is more complicated than the additive case, which is covered as a special case of Section 3. However, the same underlying principle -- the orthogonality principle -- applies for the estimation and inference on the treatment effect parameters. Estimation and inference of treatment effects in a high-dimensional setting is analysed in \citet{BCFH:Policy}.

\subsection{Treatment Effects Parameters -- a short Introduction}

In many situations researchers are asked to evaluate the effect of a policy intervention. Examples are the effectiveness of a job-related training program or the effect of a newly developed drug. We consider $n$ units or individuals, $i=1,\ldots,n$. For each individual we observe the treatment status. The treatment variable $D_i$ takes the value $1$, if the unit received (active) treatment, and $0$, if it received the control treatment. For each individual we observe the outcome for only one of the two potential treatment states. Hence, the observed outcome depends on the treatment status and is denoted by $Y_i(D_i)$. 

One important parameter of interest is the average treatment effect (ATE):
\[ \mathbb{E}[Y(1)-Y(0)] =  \mathbb{E}[Y(1)] - \mathbb{E}[Y(0)]. \]
This quantity can be interpreted as the average effect of the policy intervention.

Researchers might also be interested in the average treatment effect on the treated (ATET) given by
\[ \mathbb{E}[Y(1)-Y(0)|D=1] =  \mathbb{E}[Y(1)|D=1] - \mathbb{E}[Y(0)|D=1]. \]
This is the average treatment effect restricted to the population the treated individuals.

When treatment $D$ is randomly assigned conditional on confounding factors $X$, the ATE and ATET can be identified by regression or propensity score weighting methods.  Our identification and estimation method rely on the combination of two methods to create orthogonal estimating equations for these parameters.\footnote{It turns out that the orthogonal estimating equations are the same as doubly robust estimating equations, but emphasizing the name "doubly robust" could be confusing in the present context.}

In observational studies, the potential treatments are endogenous, i.e. jointly determined with the outcome variable.  In such cases, causal effects may be identified with the 
use of a binary instrument $Z$ that affects the treatment but is independent of the potential outcomes.  An important parameter in this case is
the local average treatment effect (LATE):
\[  \mathbb{E}[Y(1)-Y(0)| D(1) > D(0)]. \]

The random variables $D(1)$ and $D(0)$ indicate the potential participation decisions under the instrument states $1$  and $0$.
LATE is the average treatment effect for the subpopulation of compliers -- those units that respond to the change in the instrument.  Another parameter of interest is the  local average treatment effect of the treated (LATET):

\[  \mathbb{E}[Y(1)-Y(0)| D(1) > D(0), D=1], \]

which is the average effect for the subpopulation of the treated compliers. 

When the instrument $Z$ is randomly assigned conditional on confounding factors $X$, the LATE and LATET can be identified by instrumental variables regression or propensity score weighting methods.  Our identification and estimation method rely on the combination of two methods to create orthogonal estimating equations for these parameters.

\subsection{Estimation and Inference of Treatment effects}
We  consider the estimation of the effect of an endogenous binary treatment, $D$, on an outcome variable, $Y$, in a setting with very many potential control variables. In the case of endogeneity, the presence of a binary instrumental variable, $Z$, is required for the estimation of the LATE and LATET. 

When trying to estimate treatment effects, the researcher has to decide what conditioning variables to include. In the case of a non-randomly assigned treatment or instrumental variable, the researcher must select the conditioning variables so that the instrument or treatment is plausibly exogenous. Even in the case of random assignment, for a precise estimation of the policy variable selection of control variables is necessary to absorb residual variation, but overfitting should be avoided. For uniformly valid post-selection inference, \textquotedblleft orthogonal \textquotedblright estimating equations as described above are they key to the efficient estimation and valid inference. We refer to \citet{BCFH:Policy} for details.

\subsection*{R Implementation}
The package contains the functions \code{rlassoATE}, \code{rlassoATET}, \code{rlassoLATE}, and \code{rlassoLATE} to estimate the corresponding treatment effects. All functions have as arguments the outcome variable $y$, the treatment variable $d$, and the conditioning variables $x$. The functions \code{rlassoLATE}, and \code{rlassoLATE} have additionally the argument $z$ for the binary instrumental variable. For the calculation of the standard errors bootstrap methods are implemented, with options to use \code{Bayes}, \code{normal}, or \code{wild} bootstrap. The number of repetitions can be specified with the argument \code{nRep} and the default is set to $500$. By default no bootstrap standard errors are provided (\code{bootstrap="none"}). For the functions the logicals \code{intercept} and \code{post} can be specified to include an intercept and to do Post-Lasso at the selection steps. The family of treatment functions returns an object of class \code{rlassoTE} for which the methods \code{print}, \code{summary}, and \code{confint} are available.

\subsection{Application: 401(k) plan participation}
Though it is clear that 401(k) plans are widely used as vehicles for retirement saving, their effect on assets is less clear. The key problem in determining the effect of participation in 401(k) plans on
accumulated  assets  is  saver  heterogeneity  coupled  with
nonrandom selection into participation states. In particular,
it  is  generally  recognized  that  some  people  have  a  higher
preference for saving than others. Thus, it seems likely that
those individuals with the highest unobserved preference for
saving  would  be  most  likely  to  choose  to  participate  in
tax-advantaged  retirement  savings  plans  and  would  also
have  higher  savings  in  other  assets  than  individuals  with
lower unobserved saving propensity. This implies that conventional estimates that do not allow for saver heterogeneity
and  selection  of  the  participation  state  will  be  biased  upward,  tending  to  overstate  the  actual  savings  effects  of
401(k) and IRA participation.

Again, we start first with the data preparation:
\begin{knitrout}
\definecolor{shadecolor}{rgb}{0.969, 0.969, 0.969}\color{fgcolor}\begin{kframe}
\begin{alltt}
\hlkwd{data}\hlstd{(pension)}
\hlstd{y} \hlkwb{=} \hlstd{pension}\hlopt{$}\hlstd{tw}
\hlstd{d} \hlkwb{=} \hlstd{pension}\hlopt{$}\hlstd{p401}
\hlstd{z} \hlkwb{=} \hlstd{pension}\hlopt{$}\hlstd{e401}
\hlstd{X} \hlkwb{=} \hlstd{pension[,} \hlkwd{c}\hlstd{(}\hlstr{"i2"}\hlstd{,} \hlstr{"i3"}\hlstd{,} \hlstr{"i4"}\hlstd{,} \hlstr{"i5"}\hlstd{,} \hlstr{"i6"}\hlstd{,} \hlstr{"i7"}\hlstd{,} \hlstr{"a2"}\hlstd{,} \hlstr{"a3"}\hlstd{,} \hlstr{"a4"}\hlstd{,} \hlstr{"a5"}\hlstd{,} \hlstr{"fsize"}\hlstd{,}
    \hlstr{"hs"}\hlstd{,} \hlstr{"smcol"}\hlstd{,} \hlstr{"col"}\hlstd{,} \hlstr{"marr"}\hlstd{,} \hlstr{"twoearn"}\hlstd{,} \hlstr{"db"}\hlstd{,} \hlstr{"pira"}\hlstd{,} \hlstr{"hown"}\hlstd{)]}  \hlcom{# simple model}
\hlstd{xvar} \hlkwb{=} \hlkwd{c}\hlstd{(}\hlstr{"i2"}\hlstd{,} \hlstr{"i3"}\hlstd{,} \hlstr{"i4"}\hlstd{,} \hlstr{"i5"}\hlstd{,} \hlstr{"i6"}\hlstd{,} \hlstr{"i7"}\hlstd{,} \hlstr{"a2"}\hlstd{,} \hlstr{"a3"}\hlstd{,} \hlstr{"a4"}\hlstd{,} \hlstr{"a5"}\hlstd{,} \hlstr{"fsize"}\hlstd{,} \hlstr{"hs"}\hlstd{,}
    \hlstr{"smcol"}\hlstd{,} \hlstr{"col"}\hlstd{,} \hlstr{"marr"}\hlstd{,} \hlstr{"twoearn"}\hlstd{,} \hlstr{"db"}\hlstd{,} \hlstr{"pira"}\hlstd{,} \hlstr{"hown"}\hlstd{)}
\hlstd{xpart} \hlkwb{=} \hlkwd{paste}\hlstd{(xvar,} \hlkwc{collapse} \hlstd{=} \hlstr{"+"}\hlstd{)}
\hlstd{form} \hlkwb{=} \hlkwd{as.formula}\hlstd{(}\hlkwd{paste}\hlstd{(}\hlstr{"tw ~ "}\hlstd{,} \hlkwd{paste}\hlstd{(}\hlkwd{c}\hlstd{(}\hlstr{"p401"}\hlstd{, xvar),} \hlkwc{collapse} \hlstd{=} \hlstr{"+"}\hlstd{),} \hlstr{"|"}\hlstd{,} \hlkwd{paste}\hlstd{(xvar,}
    \hlkwc{collapse} \hlstd{=} \hlstr{"+"}\hlstd{)))}
\hlstd{formZ} \hlkwb{=} \hlkwd{as.formula}\hlstd{(}\hlkwd{paste}\hlstd{(}\hlstr{"tw ~ "}\hlstd{,} \hlkwd{paste}\hlstd{(}\hlkwd{c}\hlstd{(}\hlstr{"p401"}\hlstd{, xvar),} \hlkwc{collapse} \hlstd{=} \hlstr{"+"}\hlstd{),} \hlstr{"|"}\hlstd{,} \hlkwd{paste}\hlstd{(}\hlkwd{c}\hlstd{(}\hlstr{"e401"}\hlstd{,}
    \hlstd{xvar),} \hlkwc{collapse} \hlstd{=} \hlstr{"+"}\hlstd{)))}
\end{alltt}
\end{kframe}
\end{knitrout}

Now we can  compute the estimates of the target treatment effect parameters.  For ATE and ATET we report the the effect of eligibility for 401(k).

\begin{knitrout}
\definecolor{shadecolor}{rgb}{0.969, 0.969, 0.969}\color{fgcolor}\begin{kframe}
\begin{alltt}
\hlcom{# pension.ate = rlassoATE(X,d,y)}
\hlstd{pension.ate} \hlkwb{=} \hlkwd{rlassoATE}\hlstd{(form,} \hlkwc{data} \hlstd{= pension)}
\hlkwd{summary}\hlstd{(pension.ate)}
\end{alltt}
\begin{verbatim}
## Estimation and significance testing of the treatment effect 
## Type: ATE 
## Bootstrap: not applicable 
##    coeff.   se. t-value  p-value    
## TE  10490  1920   5.464 4.67e-08 ***
## ---
## Signif. codes:  
## 0 '***' 0.001 '**' 0.01 '*' 0.05 '.' 0.1 ' ' 1
\end{verbatim}
\begin{alltt}
\hlcom{# pension.atet = rlassoATET(X,d,y)}
\hlstd{pension.atet} \hlkwb{=} \hlkwd{rlassoATET}\hlstd{(form,} \hlkwc{data} \hlstd{= pension)}
\hlkwd{summary}\hlstd{(pension.atet)}
\end{alltt}
\begin{verbatim}
## Estimation and significance testing of the treatment effect 
## Type: ATET 
## Bootstrap: not applicable 
##    coeff.   se. t-value  p-value    
## TE  11810  2844   4.152 3.29e-05 ***
## ---
## Signif. codes:  
## 0 '***' 0.001 '**' 0.01 '*' 0.05 '.' 0.1 ' ' 1
\end{verbatim}
\end{kframe}
\end{knitrout}
For LATE and LATET we estimate the effect of 401(k) participation (d) with plan eligibility (z) as instrument.

\begin{knitrout}
\definecolor{shadecolor}{rgb}{0.969, 0.969, 0.969}\color{fgcolor}\begin{kframe}
\begin{alltt}
\hlstd{pension.late} \hlkwb{=} \hlkwd{rlassoLATE}\hlstd{(X, d, y, z)}
\hlcom{# pension.late = rlassoLATE(formZ, data=pension)}
\hlkwd{summary}\hlstd{(pension.late)}
\end{alltt}
\begin{verbatim}
## Estimation and significance testing of the treatment effect 
## Type: LATE 
## Bootstrap: not applicable 
##    coeff.   se. t-value  p-value    
## TE  12189  2734   4.458 8.27e-06 ***
## ---
## Signif. codes:  
## 0 '***' 0.001 '**' 0.01 '*' 0.05 '.' 0.1 ' ' 1
\end{verbatim}
\begin{alltt}
\hlstd{pension.latet} \hlkwb{=} \hlkwd{rlassoLATET}\hlstd{(X, d, y, z)}
\hlcom{# pension.latet = rlassoLATET(formZ, data=pension)}
\hlkwd{summary}\hlstd{(pension.latet)}
\end{alltt}
\begin{verbatim}
## Estimation and significance testing of the treatment effect 
## Type: LATET 
## Bootstrap: not applicable 
##    coeff.   se. t-value p-value    
## TE  12687  3590   3.534 0.00041 ***
## ---
## Signif. codes:  
## 0 '***' 0.001 '**' 0.01 '*' 0.05 '.' 0.1 ' ' 1
\end{verbatim}
\end{kframe}
\end{knitrout}

The results are summarized into a table, which is then
processed using \code{xtable} to produce the latex output:

\begin{knitrout}
\definecolor{shadecolor}{rgb}{0.969, 0.969, 0.969}\color{fgcolor}\begin{kframe}
\begin{alltt}
\hlkwd{library}\hlstd{(xtable)}
\hlstd{table} \hlkwb{=} \hlkwd{matrix}\hlstd{(}\hlnum{0}\hlstd{,} \hlnum{4}\hlstd{,} \hlnum{2}\hlstd{)}
\hlstd{table[}\hlnum{1}\hlstd{, ]} \hlkwb{=} \hlkwd{summary}\hlstd{(pension.ate)[,} \hlnum{1}\hlopt{:}\hlnum{2}\hlstd{]}
\end{alltt}
\begin{verbatim}
## Estimation and significance testing of the treatment effect 
## Type: ATE 
## Bootstrap: not applicable 
##    coeff.   se. t-value  p-value    
## TE  10490  1920   5.464 4.67e-08 ***
## ---
## Signif. codes:  
## 0 '***' 0.001 '**' 0.01 '*' 0.05 '.' 0.1 ' ' 1
\end{verbatim}
\begin{alltt}
\hlstd{table[}\hlnum{2}\hlstd{, ]} \hlkwb{=} \hlkwd{summary}\hlstd{(pension.atet)[,} \hlnum{1}\hlopt{:}\hlnum{2}\hlstd{]}
\end{alltt}
\begin{verbatim}
## Estimation and significance testing of the treatment effect 
## Type: ATET 
## Bootstrap: not applicable 
##    coeff.   se. t-value  p-value    
## TE  11810  2844   4.152 3.29e-05 ***
## ---
## Signif. codes:  
## 0 '***' 0.001 '**' 0.01 '*' 0.05 '.' 0.1 ' ' 1
\end{verbatim}
\begin{alltt}
\hlstd{table[}\hlnum{3}\hlstd{, ]} \hlkwb{=} \hlkwd{summary}\hlstd{(pension.late)[,} \hlnum{1}\hlopt{:}\hlnum{2}\hlstd{]}
\end{alltt}
\begin{verbatim}
## Estimation and significance testing of the treatment effect 
## Type: LATE 
## Bootstrap: not applicable 
##    coeff.   se. t-value  p-value    
## TE  12189  2734   4.458 8.27e-06 ***
## ---
## Signif. codes:  
## 0 '***' 0.001 '**' 0.01 '*' 0.05 '.' 0.1 ' ' 1
\end{verbatim}
\begin{alltt}
\hlstd{table[}\hlnum{4}\hlstd{, ]} \hlkwb{=} \hlkwd{summary}\hlstd{(pension.latet)[,} \hlnum{1}\hlopt{:}\hlnum{2}\hlstd{]}
\end{alltt}
\begin{verbatim}
## Estimation and significance testing of the treatment effect 
## Type: LATET 
## Bootstrap: not applicable 
##    coeff.   se. t-value p-value    
## TE  12687  3590   3.534 0.00041 ***
## ---
## Signif. codes:  
## 0 '***' 0.001 '**' 0.01 '*' 0.05 '.' 0.1 ' ' 1
\end{verbatim}
\begin{alltt}
\hlkwd{colnames}\hlstd{(table)} \hlkwb{=} \hlkwd{c}\hlstd{(}\hlstr{"Estimate"}\hlstd{,} \hlstr{"Std. Error"}\hlstd{)}
\hlkwd{rownames}\hlstd{(table)} \hlkwb{=} \hlkwd{c}\hlstd{(}\hlstr{"ATE"}\hlstd{,} \hlstr{"ATET "}\hlstd{,} \hlstr{"LATE"}\hlstd{,} \hlstr{"LATET"}\hlstd{)}
\hlstd{tab} \hlkwb{=} \hlkwd{xtable}\hlstd{(table,} \hlkwc{digits} \hlstd{=} \hlkwd{c}\hlstd{(}\hlnum{2}\hlstd{,} \hlnum{2}\hlstd{,} \hlnum{2}\hlstd{))}
\end{alltt}
\end{kframe}
\end{knitrout}

\begin{table}[ht]
\centering
\begin{tabular}{rrr}
  \hline
 & Estimate & Std. Error \\ 
  \hline
ATE & 10490.07 & 1919.99 \\ 
  ATET  & 11810.45 & 2844.33 \\ 
  LATE & 12188.66 & 2734.12 \\ 
  LATET & 12686.87 & 3590.09 \\ 
   \hline
\end{tabular}
\end{table}

Finally, we estimate a model including all interaction effects:

\begin{knitrout}
\definecolor{shadecolor}{rgb}{0.969, 0.969, 0.969}\color{fgcolor}\begin{kframe}
\begin{alltt}
\hlcom{# X = model.matrix(~ -1 + (i2 + i3 + i4 + i5 + i6 + i7 + a2 + a3 + a4 + a5 +}
\hlcom{# fsize + hs + smcol + col + marr + twoearn + db + pira + hown)^2, data =}
\hlcom{# pension) # model with interactions}
\hlstd{xvar2} \hlkwb{<-} \hlkwd{paste}\hlstd{(}\hlstr{"("}\hlstd{, xvar,} \hlstr{")^2"}\hlstd{,} \hlkwc{sep} \hlstd{=} \hlstr{""}\hlstd{)}
\hlstd{formExt} \hlkwb{=} \hlkwd{as.formula}\hlstd{(}\hlkwd{paste}\hlstd{(}\hlstr{"tw ~ "}\hlstd{,} \hlkwd{paste}\hlstd{(}\hlkwd{c}\hlstd{(}\hlstr{"p401"}\hlstd{, xvar2),} \hlkwc{collapse} \hlstd{=} \hlstr{"+"}\hlstd{),} \hlstr{"|"}\hlstd{,}
    \hlkwd{paste}\hlstd{(xvar2,} \hlkwc{collapse} \hlstd{=} \hlstr{"+"}\hlstd{)))}
\hlstd{formZExt} \hlkwb{=} \hlkwd{as.formula}\hlstd{(}\hlkwd{paste}\hlstd{(}\hlstr{"tw ~ "}\hlstd{,} \hlkwd{paste}\hlstd{(}\hlkwd{c}\hlstd{(}\hlstr{"p401"}\hlstd{, xvar2),} \hlkwc{collapse} \hlstd{=} \hlstr{"+"}\hlstd{),} \hlstr{"|"}\hlstd{,}
    \hlkwd{paste}\hlstd{(}\hlkwd{c}\hlstd{(}\hlstr{"e401"}\hlstd{, xvar2),} \hlkwc{collapse} \hlstd{=} \hlstr{"+"}\hlstd{)))}
\end{alltt}
\end{kframe}
\end{knitrout}

\begin{knitrout}
\definecolor{shadecolor}{rgb}{0.969, 0.969, 0.969}\color{fgcolor}\begin{kframe}
\begin{alltt}
\hlstd{pension.ate} \hlkwb{=} \hlkwd{rlassoATE}\hlstd{(X, z, y)}
\hlstd{pension.atet} \hlkwb{=} \hlkwd{rlassoATET}\hlstd{(X, z, y)}
\hlstd{pension.late} \hlkwb{=} \hlkwd{rlassoLATE}\hlstd{(X, d, y, z)}
\hlstd{pension.latet} \hlkwb{=} \hlkwd{rlassoLATET}\hlstd{(X, d, y, z)}
\hlcom{# pension.ate = rlassoATE(formExt, data = pension) pension.atet =}
\hlcom{# rlassoATET(formExt, data = pension) pension.late = rlassoLATE(formZExt, data =}
\hlcom{# pension) pension.latet = rlassoLATET(formZExt, data = pension)}
\hlstd{table} \hlkwb{=} \hlkwd{matrix}\hlstd{(}\hlnum{0}\hlstd{,} \hlnum{4}\hlstd{,} \hlnum{2}\hlstd{)}
\hlstd{table[}\hlnum{1}\hlstd{, ]} \hlkwb{=} \hlkwd{summary}\hlstd{(pension.ate)[,} \hlnum{1}\hlopt{:}\hlnum{2}\hlstd{]}
\hlstd{table[}\hlnum{2}\hlstd{, ]} \hlkwb{=} \hlkwd{summary}\hlstd{(pension.atet)[,} \hlnum{1}\hlopt{:}\hlnum{2}\hlstd{]}
\hlstd{table[}\hlnum{3}\hlstd{, ]} \hlkwb{=} \hlkwd{summary}\hlstd{(pension.late)[,} \hlnum{1}\hlopt{:}\hlnum{2}\hlstd{]}
\hlstd{table[}\hlnum{4}\hlstd{, ]} \hlkwb{=} \hlkwd{summary}\hlstd{(pension.latet)[,} \hlnum{1}\hlopt{:}\hlnum{2}\hlstd{]}
\hlkwd{colnames}\hlstd{(table)} \hlkwb{=} \hlkwd{c}\hlstd{(}\hlstr{"Estimate"}\hlstd{,} \hlstr{"Std. Error"}\hlstd{)}
\hlkwd{rownames}\hlstd{(table)} \hlkwb{=} \hlkwd{c}\hlstd{(}\hlstr{"ATE"}\hlstd{,} \hlstr{"ATET "}\hlstd{,} \hlstr{"LATE"}\hlstd{,} \hlstr{"LATET"}\hlstd{)}
\hlstd{tab} \hlkwb{=} \hlkwd{xtable}\hlstd{(table,} \hlkwc{digits} \hlstd{=} \hlkwd{c}\hlstd{(}\hlnum{2}\hlstd{,} \hlnum{2}\hlstd{,} \hlnum{2}\hlstd{))}
\end{alltt}
\end{kframe}
\end{knitrout}

This gives the following results:

\begin{table}[ht]
\centering
\begin{tabular}{rrr}
  \hline
 & Estimate & Std. Error \\ 
  \hline
ATE & 8449.80 & 1895.43 \\ 
  ATET  & 8938.00 & 2529.25 \\ 
  LATE & 12188.66 & 2734.12 \\ 
  LATET & 12686.87 & 3590.09 \\ 
   \hline
\end{tabular}
\end{table}

\section{The Lasso Methods for Discovery of Significant
Causes amongst Many Potential Causes, with Many Controls}
Here we consider the model
$$
\underbrace{Y_{i}}_{\mathrm{Outcome}} \ \ =  \ \ \underbrace{\sum_{l=1}^{p_1} D_{il} \alpha_\ell}_{
\mathrm{Causes}} \ \ + \ \ \underbrace{\sum_{j=1}^{p_2} 
W_{ij} \beta_j}_{\mathrm{Controls}} \ \ + \ \ \underbrace{\epsilon_i}_{\mathrm{Noise}}
$$
where the number of potential causes $p_1$ could be very large and the number of controls $p_2$ could also be very large. The causes are randomly assigned conditional on controls.

Under approximate sparsity of $ \alpha = (\alpha_l)_{l=1}^{p_1}$
and $\beta = (\beta_l)_{l=1}^{p_2}$, we can use Lasso-based method of \cite{BCK2014} for
estimating $(\alpha_l)_{l=1}^{p_1}$ and constructing a joint confidence band on  $(\alpha_l)_{l=1}^{p_1}$ and then checking which $\alpha_l$'s are significantly different from zero. The approach is based on buliding orthogonal estimating equations for each of $(\alpha_l)_{l=1}^{p_1}$, and can be interpreted as doing Frisch-Waugh procedure for each coefficient of interest, where we do partialling out via Lasso or OLS-after-Lasso.

This procedure is implemented in the R package \texttt{hdm}. Here is
an example in which $n=100$, $p_1=20$, and $p_2=20$, so that total number of regressors is $p = p_1 + p_2 = 40$.  In this example $\alpha_1 =5$ and $\beta_1 = 5$, i.e. there is only
one true cause $D_{i1}$, among the large number of causes, $D_{i1},..., D_{i20}$, and only one true control $W_{i1}$.   This example is made super-simple for clarity sake. The \cite{BCK2014} procedure, implemented by \texttt{rlasso.effects} command in R package \texttt{hdm}. 

\begin{knitrout}
\definecolor{shadecolor}{rgb}{0.969, 0.969, 0.969}\color{fgcolor}\begin{kframe}
\begin{alltt}
\hlcom{# library(hdm) library(stats)}
\hlkwd{set.seed}\hlstd{(}\hlnum{1}\hlstd{)}
\hlstd{n} \hlkwb{=} \hlnum{100}
\hlstd{p1} \hlkwb{=} \hlnum{20}
\hlstd{p2} \hlkwb{=} \hlnum{20}
\hlstd{D} \hlkwb{=} \hlkwd{matrix}\hlstd{(}\hlkwd{rnorm}\hlstd{(n} \hlopt{*} \hlstd{p1), n, p1)}  \hlcom{# Causes}
\hlstd{W} \hlkwb{=} \hlkwd{matrix}\hlstd{(}\hlkwd{rnorm}\hlstd{(n} \hlopt{*} \hlstd{p2), n, p2)}  \hlcom{# Controls}
\hlstd{X} \hlkwb{=} \hlkwd{cbind}\hlstd{(D, W)}  \hlcom{# Regressors}
\hlstd{Y} \hlkwb{=} \hlstd{D[,} \hlnum{1}\hlstd{]} \hlopt{*} \hlnum{5} \hlopt{+} \hlstd{W[,} \hlnum{1}\hlstd{]} \hlopt{*} \hlnum{5} \hlopt{+} \hlkwd{rnorm}\hlstd{(n)}  \hlcom{#Outcome}
\hlkwd{confint}\hlstd{(}\hlkwd{rlassoEffects}\hlstd{(X, Y,} \hlkwc{index} \hlstd{=} \hlkwd{c}\hlstd{(}\hlnum{1}\hlopt{:}\hlstd{p1)),} \hlkwc{joint} \hlstd{=} \hlnum{TRUE}\hlstd{)}
\end{alltt}
\begin{verbatim}
##           2.5 %     97.5 %
## V1   4.45254352 5.27634920
## V2  -0.36919897 0.35985454
## V3  -0.40022202 0.23459987
## V4  -0.30226979 0.33541589
## V5  -0.32560180 0.32529338
## V6  -0.37606144 0.34882071
## V7  -0.27299706 0.34768804
## V8  -0.09355296 0.51986258
## V9  -0.23770854 0.44138018
## V10 -0.28169032 0.30856654
## V11 -0.36118733 0.25593690
## V12 -0.36016826 0.31669947
## V13 -0.22301062 0.42568025
## V14 -0.38644116 0.44829939
## V15 -0.37829104 0.36943955
## V16 -0.31779430 0.38385080
## V17 -0.23207059 0.46982430
## V18 -0.40623680 0.08387136
## V19 -0.15173470 0.43811152
## V20 -0.25749589 0.29721605
\end{verbatim}
\begin{alltt}
\hlcom{# BCK Joint Confidence Band for Reg Coefficients 1 to 20}
\end{alltt}
\end{kframe}
\end{knitrout}
As you can see the procedure correctly tells that only
the first cause $D_{i1}$, among the large number of causes, $D_{i1},..., D_{i20}$,  is a statistically significant cause of $Y$ (see the confidence interval for variable V1).

\section{Conclusion}

We have provided an introduction to some of the capabilities of the \R package \texttt{hdm}. Inevitably, new applications will demand new features and, as the project is in its initial phase, unforeseen bugs will show up. In either case comments and suggestions of users are highly appreciated. We shall update the documentation  and the package periodically. The most current version of the \R package and its accompanying vignette will be made available at the homepage of the maintainer and \texttt{cran.r-project.org}. See the \R command \texttt{vignette()} for details on how to find
and view vignettes from within \R.

\newpage
\appendix
\section{Data Sets}
In this section we describe briefly the data sets which are contained in the package and used afterwards. They might also be of general interest either for illustrating methods or for classroom presentation.
\subsection{Pension Data}
In the United States 401(k) plans were introduced to increase
private individual  saving  for  retirement. They allow the individual to deduct  contributions  from  taxable  income  and  allow  tax-free accrual of interest on assets held within the plan (within an account).  Employers  provide  401(k)  plans,  and  employers  may  also match a certain percentage of an employee's contribution.
Because  401(k)  plans  are  provided  by  employers,  only
workers in firms offering plans are eligible for participation. This data set contains individual level information about 401(k) participation and socio-economic characteristics.

The data set can be loaded with

\begin{knitrout}
\definecolor{shadecolor}{rgb}{0.969, 0.969, 0.969}\color{fgcolor}\begin{kframe}
\begin{alltt}
\hlkwd{data}\hlstd{(pension)}
\end{alltt}
\end{kframe}
\end{knitrout}

A description of the variables and further references are given in \citet{CH401k} and at the help page, for this example called by
\begin{knitrout}
\definecolor{shadecolor}{rgb}{0.969, 0.969, 0.969}\color{fgcolor}\begin{kframe}
\begin{alltt}
\hlkwd{help}\hlstd{(pension)}
\end{alltt}
\end{kframe}
\end{knitrout}

The sample is drawn from the 1991 Survey of Income and Program Participation (SIPP) and consists of 9,915 observations. The observational units are household reference persons aged 25-64 and spouse if present. Households are included in the sample if at least one person is employed and no one is self-employed. All dollar amounts are in 1991 dollars. The 1991 SIPP reports household financial data across a
range of asset categories. These data include a variable for
whether a person works for a firm that offers a 401(k) plan.
Households in which a member works for such a
firm are classified  as  eligible  for  a  401(k).  In  addition,  the  survey also records the amount of 401(k) assets. Households with
a positive 401(k) balance are classified as participants, and eligible  households  with  a  zero  balance  are  considered
nonparticipants. Available measures of wealth in the 1991 SIPP are total wealth, net financial  assets,  and  net  non-401(k) financial  assets.  Net non-401(k)  assets  are  defined  as  the  sum  of  checking  accounts,  U.S.  saving  bonds,  other interest-earning  accounts in  banks  and  other financial institutions,  other  interest-earning assets (such as bonds held personally), stocks and mutual funds less non-mortgage debt, and IRA balances. Net financial  assets  are  net  non-401(k) financial  assets  plus 401(k) balances, and total wealth is net financial assets plus housing  equity  and  the  value  of  business,  property,  and motor vehicles.

\subsection{Growth Data}
Understanding what drives economic growth, measured in GDP, is a central question of macroeconomics. A well-known data set with information about GDP growth for many countries over a long period was collected by \citet{BarroLee1994}. This data set can be loaded by
\begin{knitrout}
\definecolor{shadecolor}{rgb}{0.969, 0.969, 0.969}\color{fgcolor}\begin{kframe}
\begin{alltt}
\hlkwd{data}\hlstd{(GrowthData)}
\end{alltt}
\end{kframe}
\end{knitrout}
This data set contains the national growth rates in GDP per capita (Outcome) for many countries with additional covariates. A very important covariate is gdpsh465, which is the initial level of per-capita GDP. For further information we refer to the help page and the references herein, in particular the online descriptions of the data set.

\subsection{Institutions and Economic Development -- Data on Settler Mortality} This data set was collected by \citet{acemoglu:colonial}  to analyse the effect of institutions on economic development. The data can be loaded with
\begin{knitrout}
\definecolor{shadecolor}{rgb}{0.969, 0.969, 0.969}\color{fgcolor}\begin{kframe}
\begin{alltt}
\hlkwd{data}\hlstd{(AJR)}
\end{alltt}
\end{kframe}
\end{knitrout}

The data set contains measurements of GDP, settler morality, an index measuring protection against expropriation risk and geographic information (latitude and continent dummies). There are  $64$ observations on 11 variables.

\subsection{Data on Eminent Domain}
Eminent domain refers to the government's taking of private property. This data set was collected to analyse the effect of the number of pro-plaintiff appellate takings decisions on economic outcome variables such as house price indices. 

The data set is loaded into \R by
\begin{knitrout}
\definecolor{shadecolor}{rgb}{0.969, 0.969, 0.969}\color{fgcolor}\begin{kframe}
\begin{alltt}
\hlkwd{data}\hlstd{(EminentDomain)}
\end{alltt}
\end{kframe}
\end{knitrout}

The data set consists of four \textquotedblleft sub data sets\textquotedblright which have the following structure:
\begin{itemize}
\item y: outcome variable, a house price index or a local GDP index,
\item d: the treatment variable, represents the number of pro-plaintiff
appellate takings decisions in federal circuit court c and year t
\item x: exogenous control variables that include a dummy variable for whether there were relevant
cases in that circuit-year, the number of takings appellate decisions, and controls for
the distribution of characteristics of federal circuit court judges in a given circuit-year
\item z: instrumental variables, here characteristics of judges serving on federal appellate panels
\end{itemize}

The four data sets differ mainly in the dependent variable, which can be repeat-sales FHFA/OFHEO house price index for metro (FHFA) and non-metro (NM) areas , the Case-Shiller home price index (CS), and state-level GDP from the Bureau of Economic Analysis. 

\subsection{BLP data}
This data set was analyzed in the seminal contribution of \cite{BLP} and stems from annual issues of the Automotive News Market Data Book. The data set inlcudes information on all models marketed during the the periord beginning 1971 and ending in 1990 cotaining 2217 model/years from 997 distinct models. A detailed description is given in \cite{BLP}, p. 868--871. The function \code{constructIV} constructs instrumental variables along the lines described and used in \cite{BLP}. The data set is loaded by
\begin{knitrout}
\definecolor{shadecolor}{rgb}{0.969, 0.969, 0.969}\color{fgcolor}\begin{kframe}
\begin{alltt}
\hlkwd{data}\hlstd{(BLP)}
\end{alltt}
\end{kframe}
\end{knitrout}
It contains information on the price (in logartihm), the market share, and car characteristics like miles per gallon, miles per dollar, horse power per weight, space and air conditioning.

\subsection{CPS data}
The CPS is a monthly U.S. household survey conducted jointly by the U.S. Census Bureau and the Bureau of Labor Statistics. The data were collected for the year 2012. The sample comprises white non-hipanic, ages 25-54, working full time full year (35+ hours per week at least 50 weeks), exclude living in group quarters, self-employed, military, agricultural, and private household sector, allocated earning, inconsistent report on earnings and employment, missing data.
It can be inspected with the command
\begin{knitrout}
\definecolor{shadecolor}{rgb}{0.969, 0.969, 0.969}\color{fgcolor}\begin{kframe}
\begin{alltt}
\hlkwd{data}\hlstd{(cps2012)}
\end{alltt}
\end{kframe}
\end{knitrout}
\newpage
\footnotesize
\bibliographystyle{econometrica}
\bibliography{mybib}

\end{document}